\documentclass[runningheads]{llncs}
\usepackage{graphicx}
\usepackage{array}
\usepackage{comment}
\usepackage{amsmath,amssymb} 
\usepackage{color}
\usepackage{multirow}

\usepackage[width=122mm,left=12mm,paperwidth=146mm,height=193mm,top=12mm,paperheight=217mm]{geometry}

\makeatother

\newcommand{\old}[1]{\iffalse{#1}\fi}

\begin{document}
\pagestyle{headings}
\mainmatter
\def\ECCVSubNumber{5473}  

\title{Bi-Directional Attention for Joint Instance and Semantic Segmentation in Point Clouds} 

\titlerunning{Bi-Directional Attention}
%
\author{Guangnan Wu\inst{1} \and Zhiyi Pan\inst{1} \and \\ 
Peng Jiang\inst{1}\thanks{Corresponding author.} \and Changhe Tu\inst{1*}}
\authorrunning{G. Wu \and Z. Pan \and P. Jiang \and C. Tu.}
%
\institute{$^1$Shandong University, China.\\
\email{\{wuguangnan1006, panzhiyi1996, sdujump, changhe.tu\}@gmail.com}}

\maketitle

\begin{abstract}
Instance segmentation in point clouds is one of the most fine-grained ways to understand the 3D scene. Due to its close relationship to semantic segmentation, many works approach these two tasks simultaneously and leverage the benefits of multi-task learning. However, most of them only considered simple strategies such as element-wise feature fusion, which may not lead to mutual promotion. In this work, we build a Bi-Directional Attention module on backbone neural networks for 3D point cloud perception, which uses similarity matrix measured from features for one task to help aggregate non-local information for the other task, avoiding the potential feature exclusion and task conflict. From comprehensive experiments and ablation studies on the S3DIS dataset and the PartNet dataset, the superiority of our method is verified. Moreover, the mechanism of how bi-directional attention module helps joint instance and semantic segmentation is also analyzed.

\end{abstract}
\section{Introduction}









Among the tasks of computer vision, instance segmentation is one of the most challenge ones which requires understand and perceive the scene in unit and instance level. Notably, the vast demands for machines to interact with real scenarios, such as robotics and autonomous driving ~\cite{nguyen20133d,ioannidou2017deep}, make the instance segmentation in the 3D scene to be the hot research topic. 

Though much progress has been made, 3D instance segmentation still lags far behind its 2D counterpart~\cite{pinheiro2015learning,he2017mask,li2017fully,dai2016instance,dai2016instance2,de2017semantic}. Unlike the 2D image, the 3D scene can be represented by many forms, such as multi-view projection images, volumes, and point clouds. Generally speaking, the form of multi-view projection images makes compromises to utilize mature techniques such as 2D CNN~\cite{su2015multi,qi2016volumetric,shi2015deeppano,guerry2017snapnet,thanh2016field} but will lose some critical information such as 3D geometry. As for representing the 3D scene as volumes~\cite{wu20153d,maturana2015voxnet,riegler2017octnet,wang2017cnn}, it simplifies the task but will lead to expensive computation and memory cost, making them impractical for complex scenarios. In contrast, point clouds could represent a 3D scene more compactly and intuitively, and thus became more popular and drew more attention recently. The proposed PointNet~\cite{qi2017pointnet} and some following works~\cite{qi2017pointnet++,huang2018recurrent,wang2019dynamic,landrieu2018large,hua2018pointwise,li2018pointcnn,ye20183d,rethage2018fully,wu2019pointconv} could process the raw point clouds directly, achieving remarkable performance on 3D classification and part segmentation tasks. The success brings the prospect for more fine-grained perception tasks in 3D point clouds, such as instance segmentation. 

Instance segmentation in point clouds requires distinguishing category and instance belonging for each point. The most direct way is to regress further the bounding box of each instance on the semantic segmentation task, such as~\cite{hou20193d,yi2019gspn,yang2019learning}. This kind of method is usually referred to as proposal-based instance segmentation, which is straightforward, but the bounding box sometimes contains multiple objects or just a part of an object, making it hard to delineate the instance precisely. For this reason, proposal-free instance segmentation is more popular. Moreover, due to the close relationship between instance segmentation and semantic segmentation, most of the recent works approach these two tasks simultaneously and use deep neural networks with two sub-branches for the two tasks, respectively~\cite{wang2019associatively,pham2019jsis3d,zhao2020jsnet}. Among them, many take feature fusion strategy letting features for one task promote the other task. However, in fact, the features of the two tasks are not completely compatible with each other. While points belong to different semantics must belong to different instances, points in the different instances are not necessarily of the different semantics. Obviously, directly concatenating or adding these two kinds of features in the model may lead to task conflict.

Actually, with simple element-wise feature fusion way such as concatenating and adding, only semantic features could always help distinguish instances in all the cases.
We will discuss the details in Sec.~\ref{sec:relate} and Sec.~\ref{sec:intro}. This situation poses a question, do we still need instance features for semantic segmentation and how to make these two tasks mutually promoted? In this work, we invested another way to incorporate features for semantic and instance segmentation. Instead of explicitly fusing features, we use similarity information implied in features for one task to assist the other task. Specifically, we first measure pair-wise similarity on semantic features to form the semantic similarity matrix, with which we propagate instance features. The propagation operation computes the response at a point as a weighted sum of the features at all points with semantic similarity as weight. Finally, the responses are further concatenated to the original instance features for instance segmentation. The same steps are also conducted in another direction that computing instance similarity matrix to propagate semantic features for semantic segmentation. The propagation operation could aggregate non-local information and is also referred to as attention~\cite{wang2018non,vaswani2017attention,zhao2018psanet,fu2019dual}. Therefore, we name this kind of module as Bi-Directional Attention and call our networks as BAN.

\begin{figure}[htbp]
\centering
\includegraphics[scale=0.3]{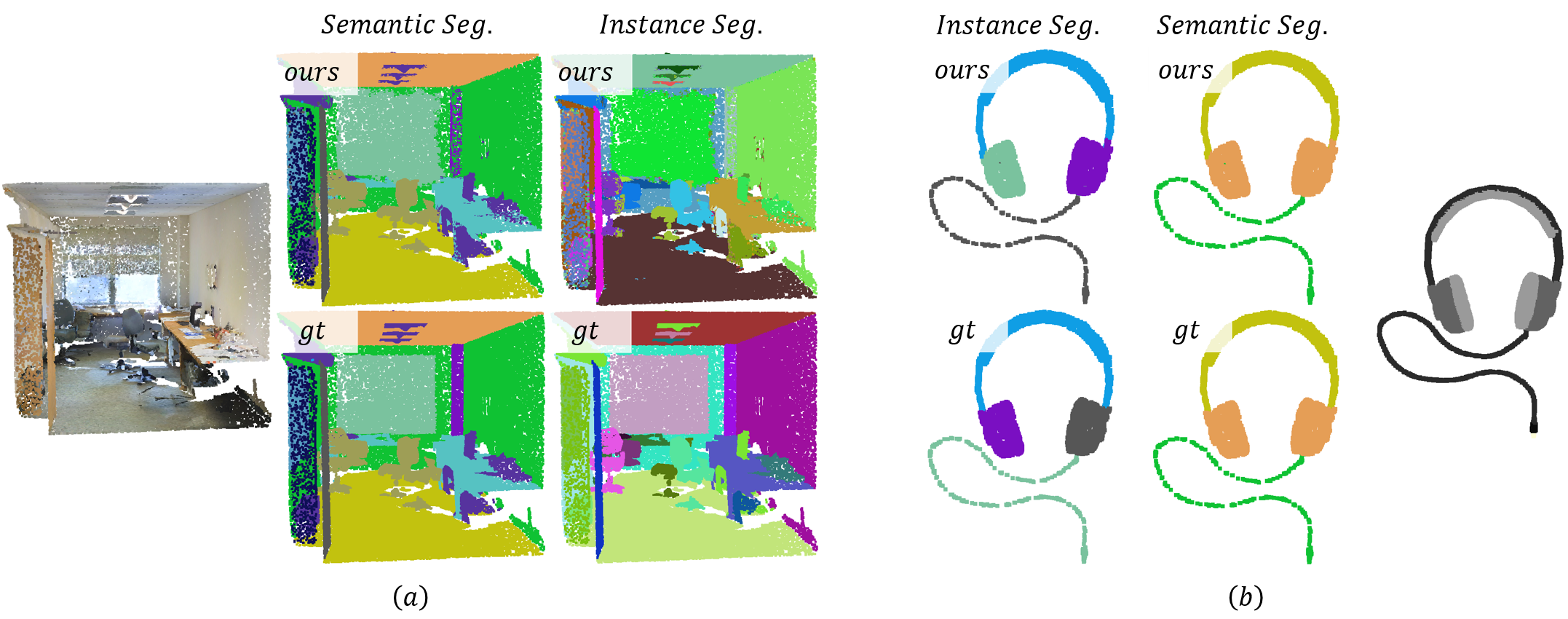}
\caption{Instance and semantic segmentation in point clouds using BAN. (a) Results on the S3DIS dataset, (b) Results on the PartNet dataset.}
\label{fig:fig1}
\end{figure}

The help of aggregating non-local information lies in the following aspects. First, for attention applied to instance features for instance segmentation, semantic similarity matrix would help push instance features belonging to the different semantic apart. Though it will also pull instance features belonging to the same semantic together, the concatenated original instance features could still guarantee the difference distinguishable. Second, for attention applied on semantic features for semantic segmentation, instance similarity matrix would let semantic within each instance more consistent, thus improve the detail delineation. In addition to the positive effects when using bi-directional attention in a forward manner, the attention operation will also be good for back-propagating uniform gradients within the same semantic or instance.
Consequently, our Bi-Directional Attention module could aggregate the features more properly and avoid potential task conflict. We compare our BAN to state-of-the-art methods on prevalent 3D point cloud datasets, including S3DIS\cite{armeni20163d} and PartNet~\cite{mo2019partnet}. We demonstrate two instance and semantic segmentation results in Fig.~\ref{fig:fig1}.
In experiments, our method demonstrates consistent superiority according to most of the evaluation metrics. Moreover, we conduct detailed ablation and mechanism studies, which suggests that the similarity matrices truly reflect the required pair-wise semantic and instance similarities. With attention operations from two directions together sequentially, we can reach the best performance. Our code has been open sourced.






\section{Related Works}\label{sec:relate}
In this section, we will revisit some most relevant works of instance segmentation in point clouds. These works could be divided into two types in general, proposal-based and proposal-free.
\subsection{Proposal-based methods}
Most proposal-based methods in point clouds also follow the scheme of Mask R-CNN~\cite{he2017mask} in 2D images, which forms instance segmentation as joint object bounding box regression and semantic segmentation. 3D-SIS~\cite{hou20193d} and GSPN~\cite{yi2019gspn} rely on anchors and two-stage training, which will spend additional time to prune the dense object proposals.
BoNet~\cite{yang2019learning} directly regresses bounding box prediction without anchors. However, only global features are used to regress rough instance boxes.
\subsection{Proposal-free methods}
Proposal-free methods directly produce representations to estimate the semantic categories and cluster the instance groups.
SGPN~\cite{wang2018sgpn} learns a similarity matrix to group instance and treats semantic segmentation as a standalone task. 3D-BEVIS~\cite{elich20193d} gets additional instance feature from birds-eye-view, but still considers semantic segmentation independent of instance segmentation. 

In view of the close relationship between instance and semantic segmentation, many works started to study how to incorporate the features of two tasks efficiently for mutual benefits. 
JSIS3D~\cite{pham2019jsis3d} uses multi-value conditional random field to fuse semantic and instance, but it requires some approximation to optimize. ASIS~\cite{wang2019associatively} fuses semantic features to instance features by element-wise add to help distinguish instances of the different semantics. Besides, the KNN is used to assemble more instance features from the neighborhood to each point and make the assembled feature more robust, but it is non-differentiable and will break the back-propagation chain. The use of KNN in this work could be considered as proto non-local operation.

The most recent work JSNet~\cite{zhao2020jsnet} fuses semantic and instance features to each other by simple aggregation strategies such as element-wise add and concatenate operations. In this way, the problem can be formalized as the following equations:
\begin{equation}
\begin{aligned}
    \mathcal{F}(\alpha(S_a, I_a)) \rightarrow C_a, \mathcal{F}(\alpha(S_b, I_b)) \rightarrow C_b,\\
    \mathcal{H}(\alpha(S_a, I_a)) \rightarrow G_a, \mathcal{H}(\alpha(S_b, I_b)) \rightarrow G_b,\\
\end{aligned}
\label{eq:1}
\end{equation}
where $S_i$ and $I_i$ represent semantic and instance features of point $i$ respectively, and $C_i$ and $G_i$ are the semantic category and instance group of point $i$. $\alpha$ is some simple feature aggregating method. We use $\mathcal{F}$ and $\mathcal{H}$ to represent mapping functions for semantic and instance segmentation, respectively.

Ideally, there are three cases for two points $a$ and $b$: (1) $C_a$=$C_b$ and $G_a$$\neq$$G_b$; (2) $C_a$=$C_b$ and $G_a$=$G_b$; (3) $C_a$$\neq$$C_b$ and $G_a$$\neq$$G_b$. 
In the first case, for semantic segmentation $\mathcal{F}$, aggregating $S$ and $I$ by $\alpha$ will make responses $\alpha(S_a, I_a)$ and $\alpha(S_b, I_b)$ far away. Thus $C_a$ and $C_b$ are hard to keep consistent, which is contrary to the case setting.
In the second case, both $\mathcal{F}$ and $\mathcal{H}$ could get promoted by aggregating features of the same instance by $\alpha$.
The third case will not be considered when aggregating feature, because $a$ and $b$ are not relevant in either semantic or instance.
So, with the simple aggregation strategy adopted by JSNet~\cite{zhao2020jsnet}, there is a potential risk of task conflict in some specific cases.

In summary, though the non-local operation and feature aggregation strategy demonstrated certain advantages, the current implementation has some crucial problems. Considering this, we invest a proper non-local feature aggregation method in this work.
\section{Methodology and Implementation}\label{sec:intro}
\subsection{Methodology}
Directly adding or concatenating semantic and instance features for semantic segmentation may raise some problems as discussed in Sec.~\ref{sec:relate}. However, the similarity information implied in the instance features would help semantic segmentation without any harm. Here we propose a way to use similarity information.

We adjust the point's semantic feature as the weighted sum of semantic features of points belong to the same instance (with similar instance features). This way would make the semantic features robust and consistent within each instance, which will promote the details delineation. To enable this function and take advantage of similar information in the instance features, we design the aggregation operation as:
\begin{equation}
\begin{aligned}
    &\alpha(X,Y) = \{Pg(Y),Y\},\\
    &P = softmax(\theta(X)\phi(X)^T),\\
\end{aligned}
\label{eq:eq1}
\end{equation}
where $X$ and $Y$ represent two kinds of features of size $N\times{N_X}$ and $N\times{N_Y}$ respectively ($N$ is point number and $N_i$ is number of channels for feature $i$).
$\theta$, $\phi$ and $g$ are functions to re-weighted sum values in feature dimension with learned weights. We measure similarities by inner-product of $\theta(X)$ and $\phi(X)$, which results into a matrix of size $N\times{N}$. We further apply $softmax$ on each row to get transition matrix which is our final similarity matrix $P$.

When $X$ is instance features, and $Y$ is semantic features, this operation propagates semantic features to other points by instance similarity matrix, the adjusted semantic features $Pg(Y)$ will be more uniform in each instance than the original $Y$. Since there is no explicit element-wise feature adding or concatenating, using the final aggregation result $\alpha(X, Y)$ for semantic segmentation will not have the problem mentioned in the last section. Besides, this aggregation operation has the non-local characteristic naturally. For these reasons, we will also use it to fuse semantic features for instance segmentation. In other words, we will conduct another aggregation operation with $X$ as semantic features and $Y$ as instance features for instance segmentation. Consequently, in our method, we have two attentions with different data flow directions, which we name the Bi-Directional Attention module.

\begin{figure}[htbp]
\centering
\includegraphics[scale=0.4]{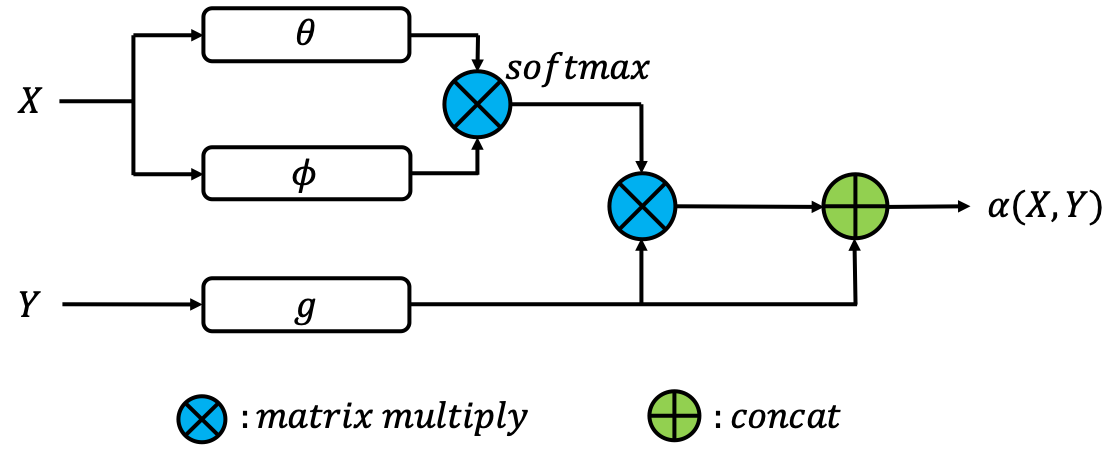}
\caption{Attention operation.}
\label{fig:attention}
\end{figure}

It is worth noting that the above-defined aggregation operation has a similar form as attention operation in~\cite{wang2018non}, but ours has two kinds of inputs for joint instance and segmentation in point clouds. The architecture of our attention (aggregation) operation is illustrated in Fig.~\ref{fig:attention}.

\subsection{Implementation}\label{sec:implet}
\subsubsection{Networks}\label{sec:networks}
\begin{figure}[htbp]
\centering
\includegraphics[scale=0.2]{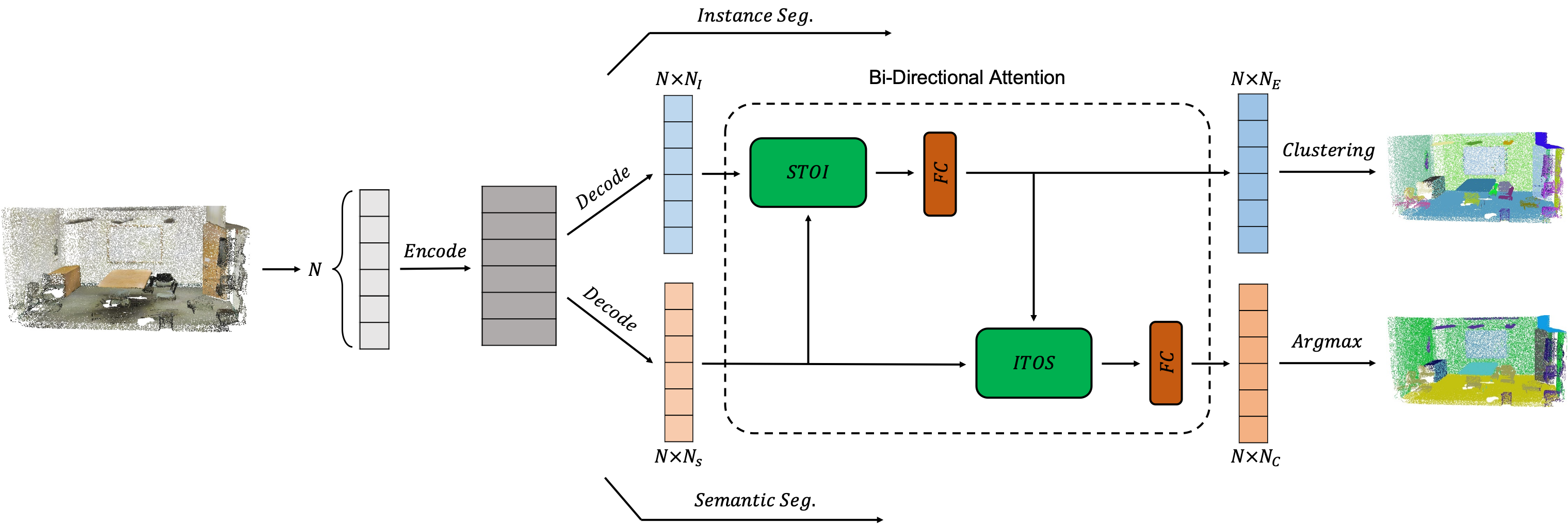}
\caption{The pipeline of proposed Bi-Directional Attention Networks (BAN).}
\label{fig:full_pipeline}
\end{figure}

By connecting the Bi-Directional Attention module to the end of the feature extracting backbone, we have the Bi-Directional Attention networks (BAN), which uses two attention operations to achieve information transmission and aggregation between instance branch and semantic branch. The full pipeline of our networks is illustrated in Fig.~\ref{fig:full_pipeline}.

Our BAN is composed of a shared encoder, and two parallel decoders to produce representations for estimating the semantic categories and clustering the instance groups. 
Specifically, our backbone is PointNet++~\cite{qi2017pointnet++}. Given input point clouds of size $N$, the backbone first extracts and encodes them into feature matrix which further decoded to semantic feature matrix $S$ of size $N\times{N_S}$ and instance feature matrix $I$ of $N\times{N_I}$.

The Bi-Directional Attention module takes these two feature matrices as input and will conduct two attention operations as defined by Eq.~\ref{eq:eq1}. We name the attention operation that computes semantic similarity matrix applied to instance features for instance segmentation as STOI, and attention operation that computes instance similarity matrix applied to semantic features for semantic segmentation as ITOS. The output of STOI is further passed to some simple fully connected layers (FC) to produces instance embedding space (of size $N\times{N_E}$), while the output of ITOS is further passed to some simple fully connected layers (FC) to give semantic prediction (of size $N\times{N_C}$). To get the instance groups, we cluster the produced instance embedding space by mean-shift method~\cite{cheng1995mean}.

There are three kinds of sequences to conduct STOI and ITOS, and they are STOI first, ITOS first, and simultaneously. Here we use STOI first because we will use pixel-level regression loss for semantic segmentation and discriminative loss for instance segmentation, and we believe semantic features will converge faster than instance features. So, semantic features will give instance segmentation task more help at the beginning. This assumption will be verified in our ablation study in Sec.~\ref{sec:exp}.

\subsubsection{Loss Function}
Our loss function $\mathcal{L}$ has two parts, semantic segmentation loss $\mathcal{L}_{sem}$ and instance segmentation loss $\mathcal{L}_{ins}$, and optimized at the same time:
\begin{equation}
    \mathcal{L} = \mathcal{L}_{sem} + \mathcal{L}_{ins}.
\end{equation}

We use classical cross-entropy loss for $\mathcal{L}_{sem}$, and choose discriminative loss function for 2D images in~\cite{de2017semantic} as $\mathcal{L}_{ins}$. 
The discriminative loss has been extended to 3D point clouds and used by many works~\cite{wang2019associatively,pham2019jsis3d,zhao2020jsnet}.
$\mathcal{L}_{ins}$ will penalize the grouping of the points across different instances and bring the points belonging to the same instance closer in the embedding space. For the detailed definition of loss function, please check the supplementary.

\subsubsection{Derivative Analysis}\label{sec:grad}
The above sections have explained how our Bi-Directional Attention module gives help in a forward manner. Here we further analyze the back-propagation of proposed Eq.~\ref{eq:eq1}.
To simplify the problem, we first give a simple version of Eq.~\ref{eq:eq1} without softmax, re-weight functions, and concatenation of original features:
\begin{equation}
\begin{aligned}
    Z = XX^TY.
\end{aligned}
\label{eq:eq2}
\end{equation}
where $Z$ is the output of simplified attention operation.
In this case, the derivatives with respect to feature $X$ and $Y$ are:

\begin{flalign}
	vec(d\mathcal{L}) &= (\frac{\partial{\mathcal{L}}}{\partial{Z}})^Tvec(dZ)\nonumber\\
			&= (\frac{\partial{\mathcal{L}}}{\partial{Z}})^T[vec(dXX^TY) + vec(XdX^TY)\nonumber]\\
			&= (\frac{\partial{\mathcal{L}}}{\partial{Z}})^T[(X^TY)^T \otimes E_N + (Y^T \otimes X)K_{NN_{X}}]vec(dX)\\
	\frac{\partial{\mathcal{L}}}{\partial{X}} &= [(X^TY) \otimes E_{N} + K_{NN_X}(Y \otimes X^T)]\frac{\partial{\mathcal{L}}}{\partial{Z}}\nonumber\\\nonumber\\
	vec(d\mathcal{L}) &= (\frac{\partial{\mathcal{L}}}{\partial{Z}})^Tvec(dZ)\nonumber\\
			&= (\frac{\partial{\mathcal{L}}}{\partial{Z}})^Tvec(XX^TdY)\nonumber\\
			&= (\frac{\partial{\mathcal{L}}}{\partial{Z}})^T(E_{N_Y} \otimes XX^T)vec(dY)\\
	\frac{\partial{\mathcal{L}}}{\partial{Y}} &= (E_{N_Y} \otimes XX^T)\frac{\partial{\mathcal{L}}}{\partial{Z}}\nonumber\\\nonumber
\end{flalign}
where $vec()$ means matrix vectorization and $\otimes$ represents Kronecker Product,
$E$ is identity matrix and $K$ is commutation matrix.

It can be seen, the similarity matrices also appear in $\frac{\partial{\mathcal{L}}}{\partial{X}}$ and $\frac{\partial{\mathcal{L}}}{\partial{Y}}$. As for $XX^T$ in $\frac{\partial{\mathcal{L}}}{\partial{Y}}$, it will make the gradients uniform and robust within a similar region defined by $X$ (semantic or instance), thus help optimization. As for $X^TY$, it computes similarities between different features of $X$ and $Y$ other than points and provides another crucial information to extract robust and useful gradients.

In summary, the proposed Bi-Directional Attention module not only help joint instance and semantic segmentation by transmitting and aggregating information between instance features and semantic features, and also be good for back-propagating uniform and robust gradients.

\section{Experiments}\label{sec:exp}
\subsection{Experiments setting}
\subsubsection{Datasets}
We study and evaluate our method on prevalent used two datasets.
\emph{Stanford 3D Indoor Semantics Dataset (S3DIS)~\cite{armeni20163d}} contains 3D scans in 6 areas including 271 rooms. Each scanned 3D point is associated with an instance label and a semantic label from 13 categories. \emph{PartNet~\cite{mo2019partnet}} contains 573,585 fine-grained part instances with annotations and has 24 object categories.



\subsubsection{Evaluation Metrics}
For semantic segmentation, we compare our BAN with others by overall accuracy(oAcc), mean accuracy (nAcc), and mean IoU (mIoU).
As for instance segmentation, coverage (Cov) and weighted coverage (WCov) \cite{ren2017end,liu2017sgn,zhuo2017indoor} are adopted.
Cov and Wcov are defined as:
\begin{equation}
    Cov(\mathcal{G,O}) = \sum\limits_{i=1}^{|\mathcal{G}|}\frac{1}{|\mathcal{G}|}\max\limits_{j}IoU(r_{i}^{G},r_{j}^{O})
\end{equation}
\begin{equation}
    WCov(\mathcal{G,O}) = \sum\limits_{i=1}^{|\mathcal{G}|}\frac{1}{|\mathcal{G}|}\omega_{i}\max\limits_{j}IoU(r_{i}^{G},r_{j}^{O})
\end{equation}
\begin{equation}
    \omega_{i} = \frac{|r_{i}^{G}|}{\sum_{k}|r_{k}^{G}|}
\end{equation}\\
where ground-truth region is denoted as $\mathcal{G}$ and predicted regions is denoted as $\mathcal{O}$, $|r_{i}^{G}|$ is the number of points in ground-truth region $i$. Besides, the classical metrics mean precision (mPrec), and mean recall (mRec) with IoU threshold $0.5$ are also reported.

\subsubsection{Training and Testing Details}
To optimize our Bi-Directional Attention Networks (BAN), we use Adam optimizer~\cite{adam} with batch size $12$ and set initial learning rate as $0.001$ following the “divided by $2$ every $300k$ iterations” learning rate policy. During training, we carry out $100$ epochs in total and use the default parameter setting in~\cite{de2017semantic} for $\mathcal{L}_{ins}$. At test time, bandwidth is set to $0.6$ for mean-shift clustering. BlockMerging algorithm proposed by SGPN~\cite{wang2018sgpn} is used to merge instances from different blocks.

For S3DIS~\cite{armeni20163d}, we carry out training and testing with the $6$-fold cross-validation and split the rooms into $1m$ $\times$ $1m$ overlapped blocks (each containing $4096$ points) on the ground plane, as used in~\cite{qi2017pointnet}. 
While for PartNet~\cite{mo2019partnet}, as~\cite{wang2018sgpn}, we train and test on each object category separately and report the evaluation results as the mean of metric values over all the objects.

\begin{table}
    \centering
    \setlength{\belowcaptionskip}{5pt}
    \caption{Instance segmentation results on S3DIS dataset.}
    \begin{tabular}{c|c|c|c|c|c}
        \hline \hline
        Method&Backbone&mCov&mWCov&mPrec&mRec\\
        \hline\hline
        \multicolumn{6}{c}{Test on 6-fold cross-validation}\\
        \hline
        PointNet&PointNet&43.0&46.3&50.6&39.2\\
        \hline
        PointNet++&PointNet++&49.6&53.4&62.7&45.8\\
        \hline
        SGPN&PointNet&37.9&40.8&38.2&31.2\\
        \hline
        ASIS&PointNet++&51.2&55.1&63.6&47.5\\
        \hline
        BoNet&PointNet++&46.0&50.2&\textbf{65.6}&47.6\\
        \hline
        Ours&PointNet++&\textbf{52.1}&\textbf{56.2}&63.4&\textbf{51.0}\\
        \hline
    \end{tabular}
    \label{tab:ins}
\end{table}
\begin{table}
    \centering
    \setlength{\belowcaptionskip}{5pt}
    \caption{Semantic segmentation results on S3DIS dataset.}
    \begin{tabular}{c|c|c|c|c}
        \hline \hline
        Method&Backbone&mAcc&mIoU&oAcc\\
        \hline\hline
        \multicolumn{5}{c}{Test on 6-fold cross-validation}\\
        \hline
        PointNet&PointNet&60.7&49.5&80.4\\
        \hline
        PointNet++&PointNet++&69.0&58.2&85.9\\
        \hline
        ASIS&PointNet++&70.1&59.3&86.2\\
        \hline
        Ours&PointNet++&\textbf{71.7}&\textbf{60.8}&\textbf{87.0}\\
        \hline
    \end{tabular}
    \label{tab:sem}
\end{table}

\subsection{S3DIS Results}
In this section, we will compare our method (BAN) with other state-of-the-art methods, and the reported metric values are either from their papers or implemented and evaluated by ourselves when not available.

\subsubsection{Instance segmentation}
In Tab.~\ref{tab:ins}, six methods are compared, including PointNet\cite{qi2017pointnet}, PointNet++\cite{qi2017pointnet++},  SGPN\cite{wang2018sgpn}, ASIS\cite{wang2019associatively}, BoNet\cite{yang2019learning} and our BAN. It's worth to note that, PointNet++ has the same architecture and settings as ours except the Bi-Directional Attention module, and thus can be treated as baseline. PointNet is similar to PointNet++ except the backbone. It can be seen, our BAN outperforms baseline (PointNet++) on all the metrics, and demonstrates significant superiority compared with others.

The more detailed comparison by $WCov$ on each of $13$ categories are shown in Tab.~\ref{tab:per class}. Ours get the highest score on most of the categories.

\subsubsection{Semantic segmentation}
Since SGPN\cite{wang2018sgpn} and BoNet\cite{yang2019learning} do not provide semantic segmentation results. For semantic segmentation, we only compare PointNet\cite{qi2017pointnet}, PointNet++\cite{qi2017pointnet++} and ASIS\cite{wang2019associatively}.

The evaluation results are shown in Tab.~\ref{tab:sem}, from mAcc, mIoU, and oAcc, our method achieves the best performance consistently. Evaluations on all the $13$ semantic categories by $mIoU$ are listed in Tab.~\ref{tab:per class}, and we get the best performance on most of that.

\begin{table}[htbp]
	\centering
	\setlength{\belowcaptionskip}{5pt}
	\caption{Per class results on S3DIS dataset}
	\begin{tabular}{ccccccccc}
	\hline\hline
	&mean&ceiling&floor&wall&beam&column&window&\multirow{2}*{method}\\
	&door&table&chair&sofa&bookcase&board&clutter&\\
	\hline\hline
	\multirow{6}*{WCov}&54.7&82.1&\textbf{78.3}&69.2&40.0&18.4&57.9&\multirow{2}*{ASIS}\\
	&\textbf{59.1}&58.0&63.1&36.2&\textbf{44.3}&54.5&50.4&\\
	\cline{2-9}
	&50.2&78.3&70.5&68.2&38.3&15.4&55.3&\multirow{2}*{BoNet}\\
	&56.2&51.9&\textbf{67.2}&24.5&36.7&42.3&47.6&\\
	\cline{2-9}
	&\textbf{56.2}&\textbf{82.7}&76.8&\textbf{69.7}&\textbf{44.4}&\textbf{20.3}&\textbf{60.9}&\multirow{2}*{Our}\\
	&58.4&\textbf{59.2}&62.9&\textbf{41.2}&\textbf{44.3}&\textbf{56.2}&\textbf{51.9}&\\
	\hline
	\multirow{4}*{mIoU}&59.7&\textbf{93.9}&\textbf{95.7}&74.9&36.1&30.0&53.4&\multirow{2}*{ASIS}\\
	&63.3&63.0&\textbf{70.6}&36.8&50.1&49.9&\textbf{58.2}&\\
	\cline{2-9}
	&\textbf{60.8}&\textbf{93.9}&94.2&\textbf{77.0}&\textbf{38.0}&\textbf{32.6}&\textbf{54.9}&\multirow{2}*{Our}\\
	&\textbf{64.5}&\textbf{65.8}&68.2&\textbf{38.6}&\textbf{52.2}&\textbf{52.3}&\textbf{58.2}&\\
	\hline
	\end{tabular}
	\label{tab:per class}
\end{table}

\subsubsection{Visual Comparison}
\begin{figure}[htbp]
\centering
\begin{tabular}{p{2cm}<{\centering} p{2cm}<{\centering} p{2cm}<{\centering} p{2cm}<{\centering} p{2cm}<{\centering} p{2cm}<{\centering}}
	\includegraphics[scale=0.08]{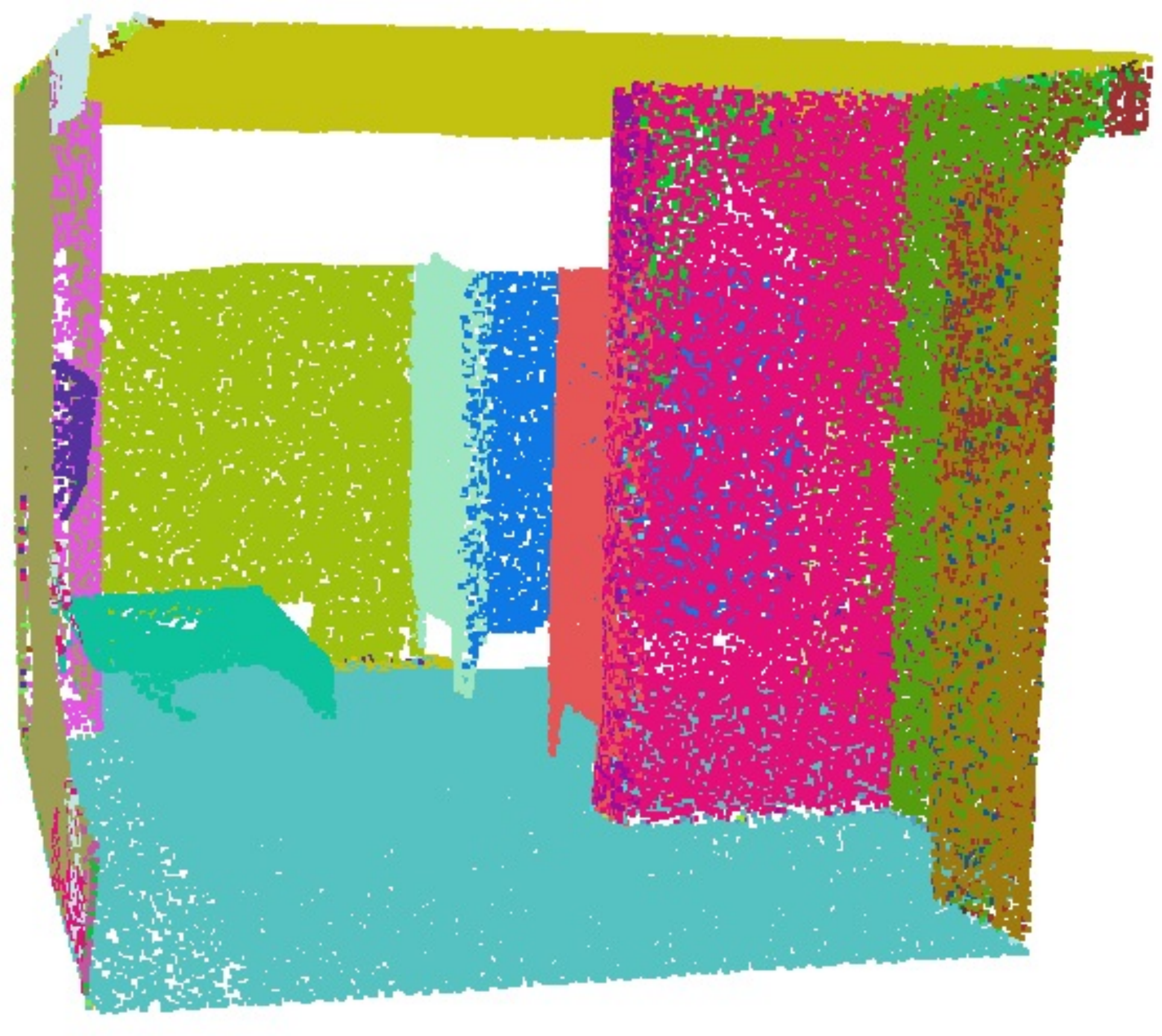}&\includegraphics[scale=0.08]{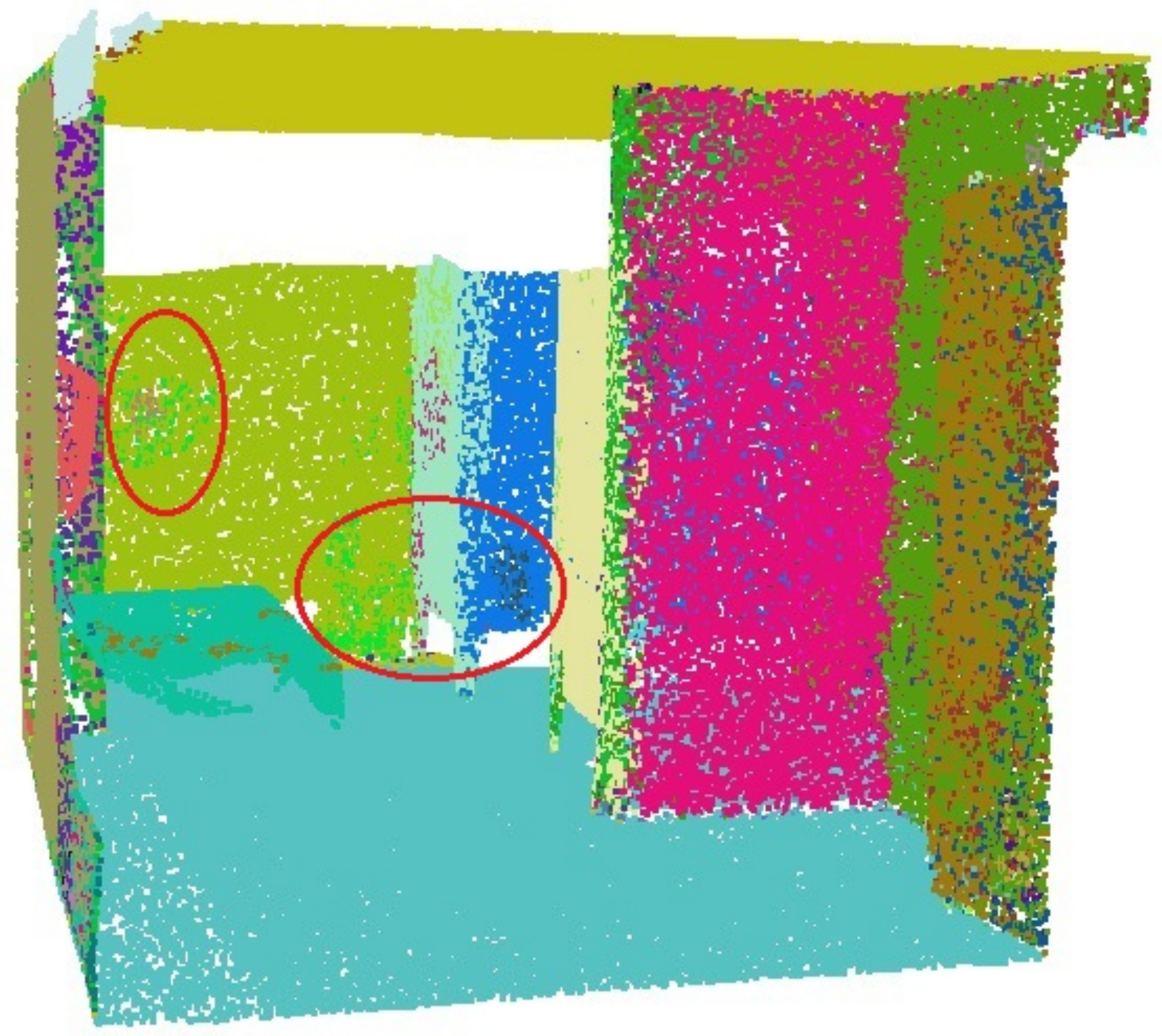}&\includegraphics[scale=0.08]       {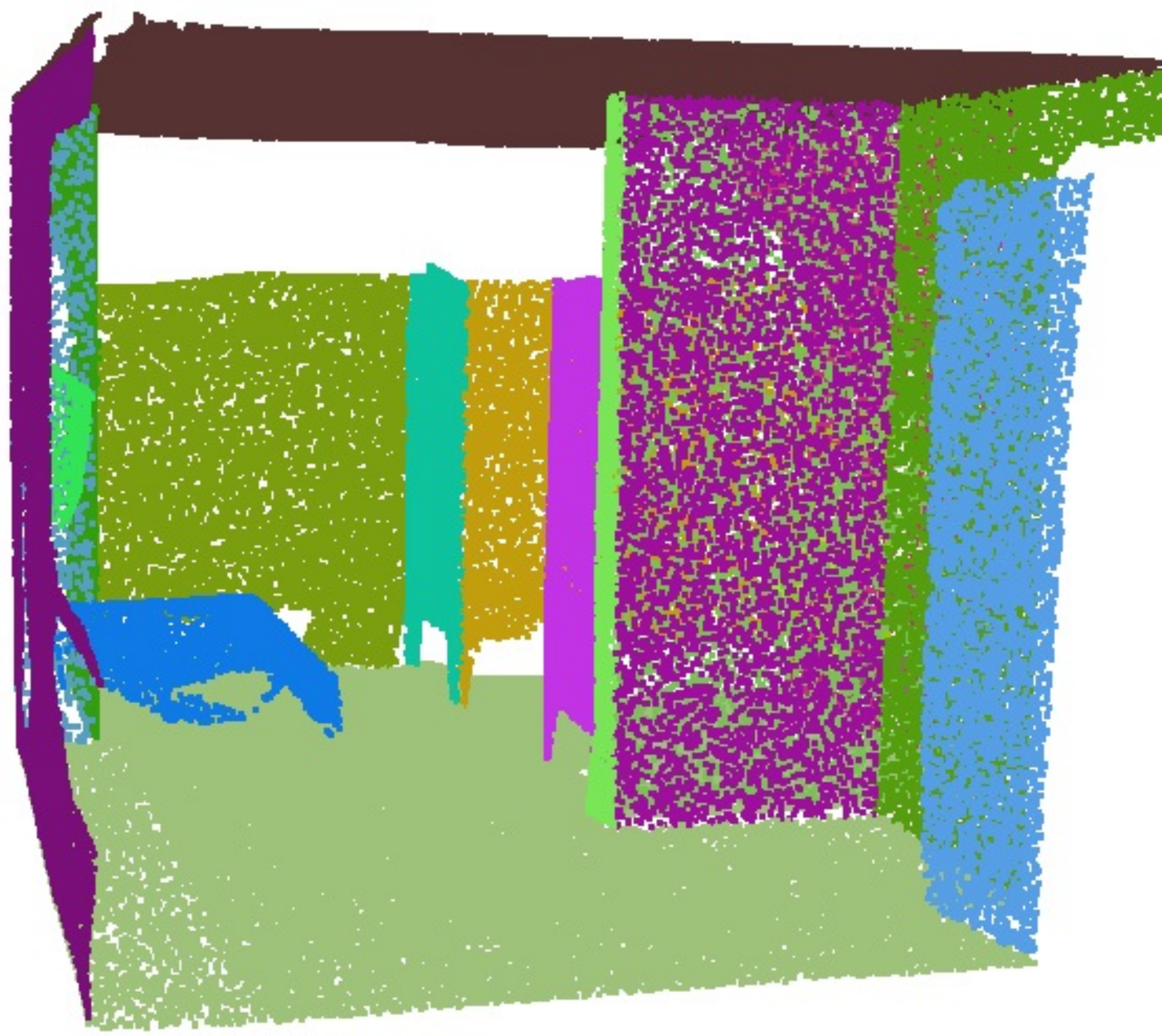}&\includegraphics[scale=0.08]{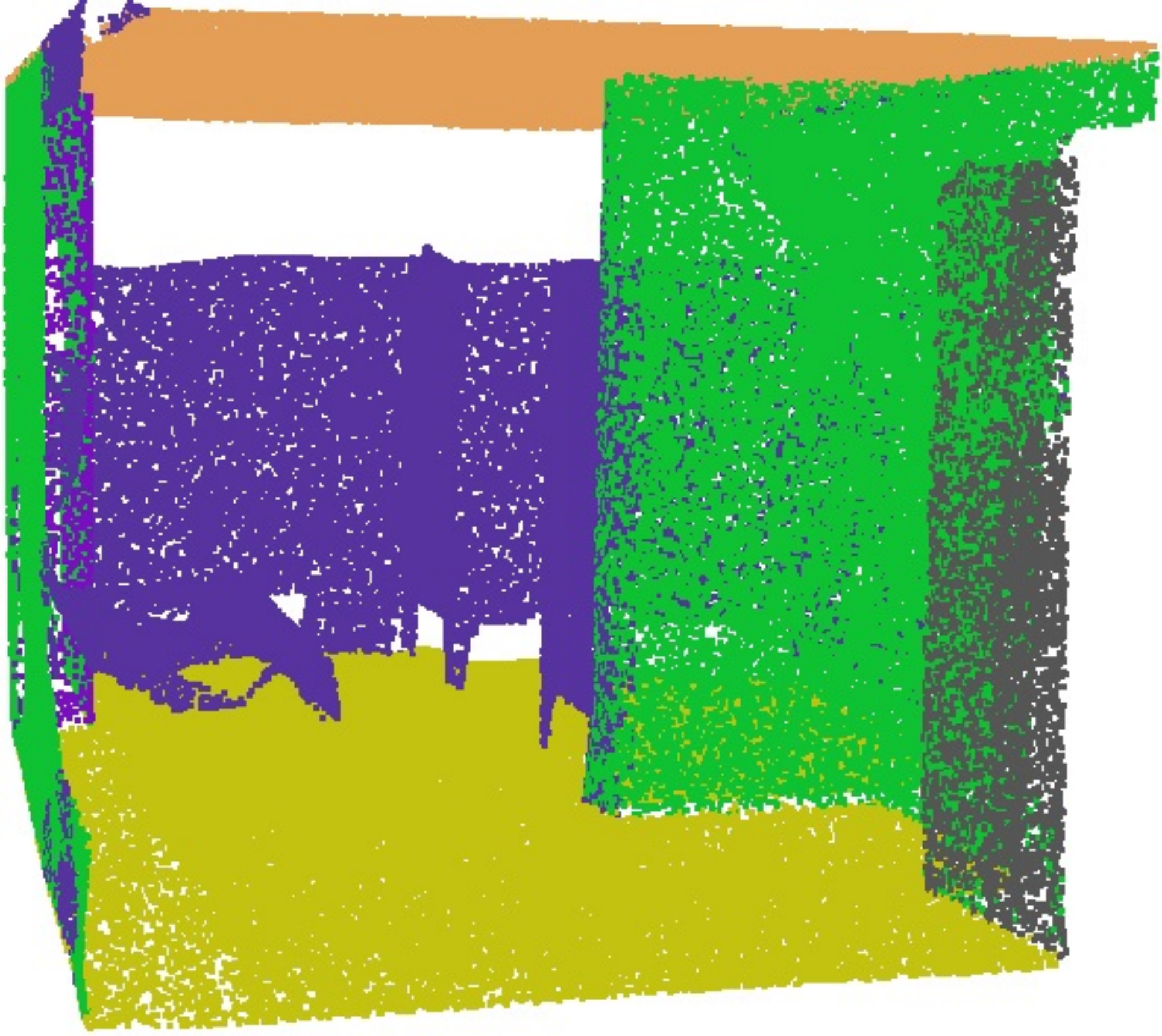}&\includegraphics[scale=0.08]{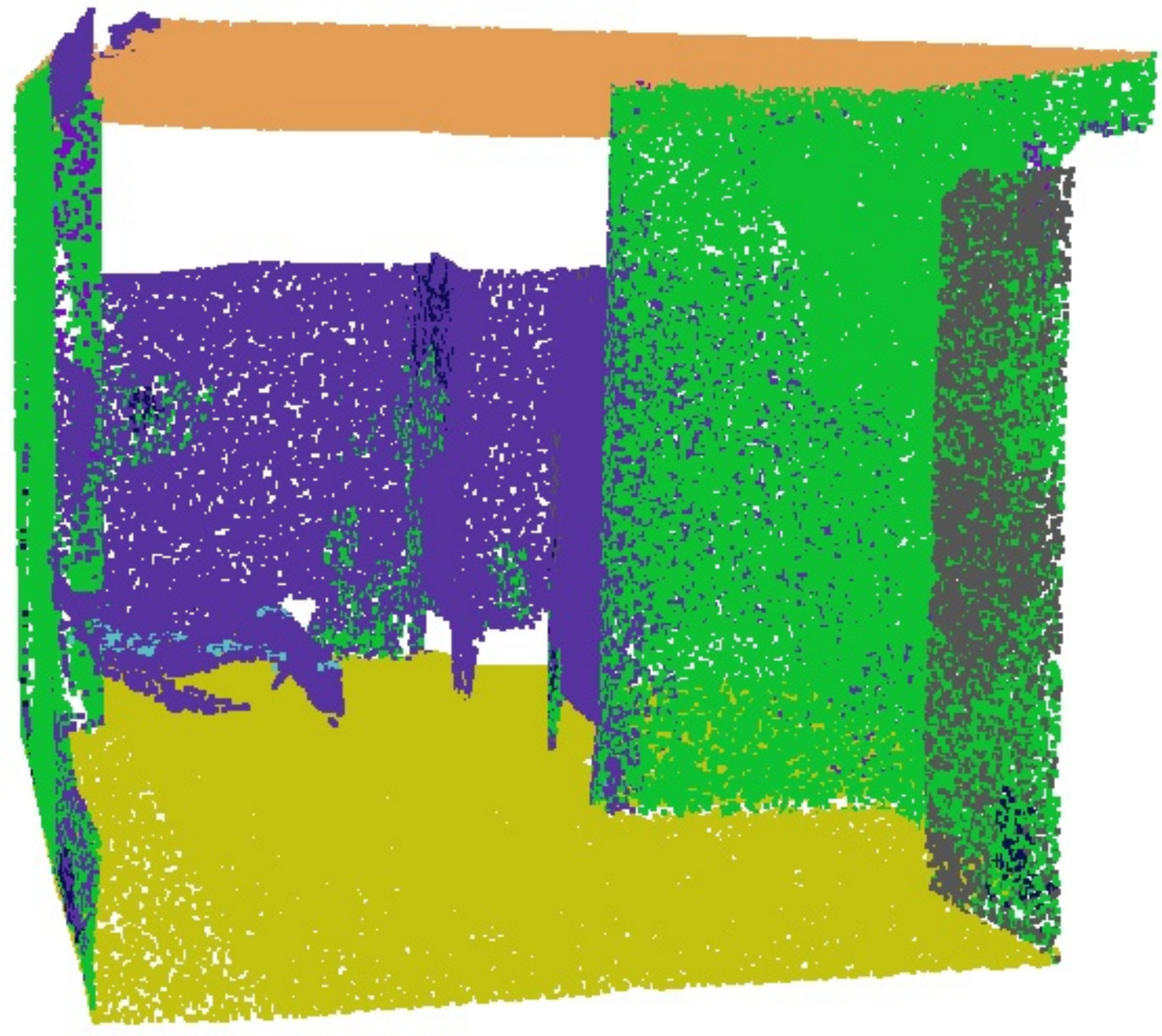}&\includegraphics[scale=0.08]{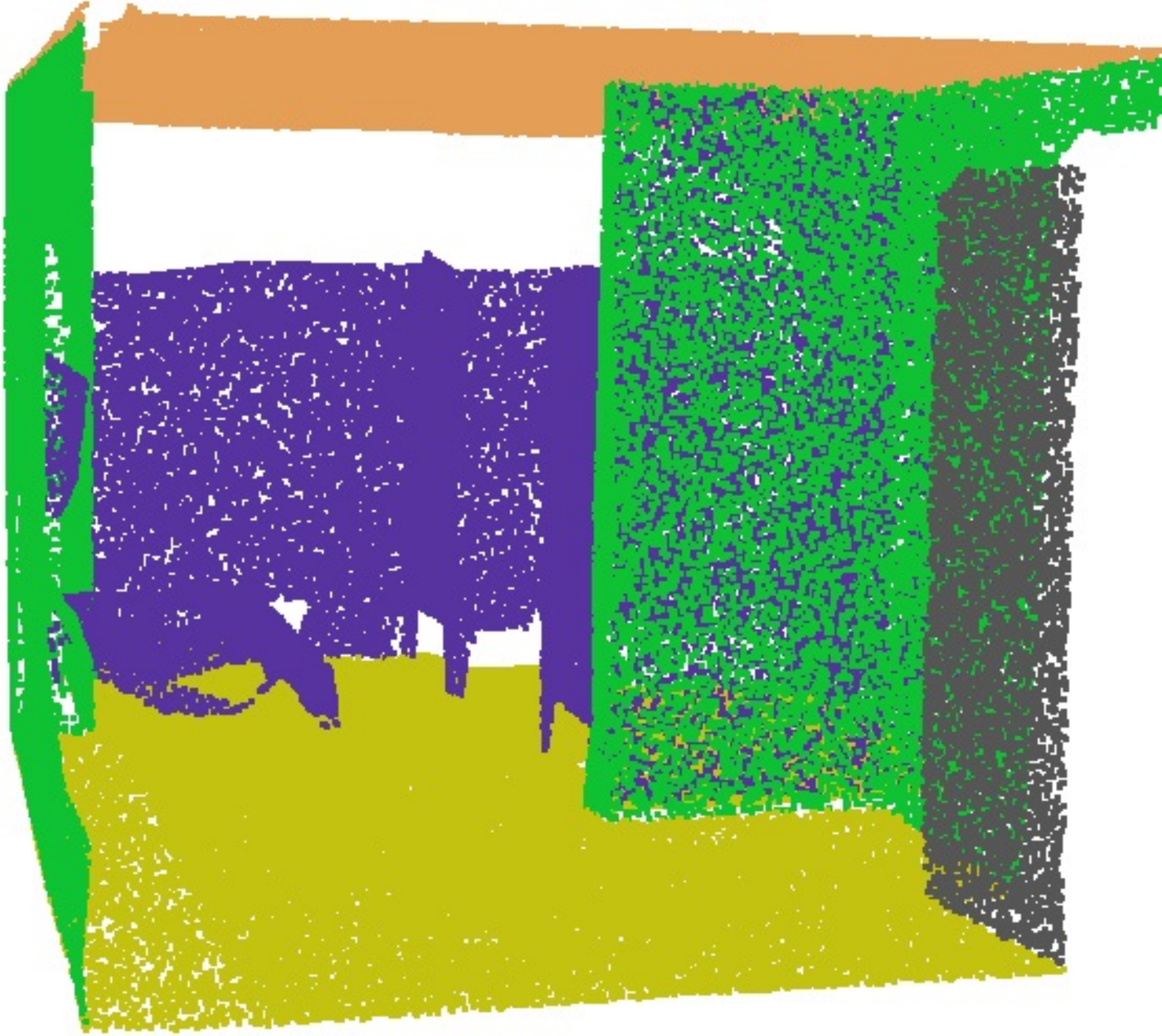}\\\\
	\includegraphics[scale=0.06]{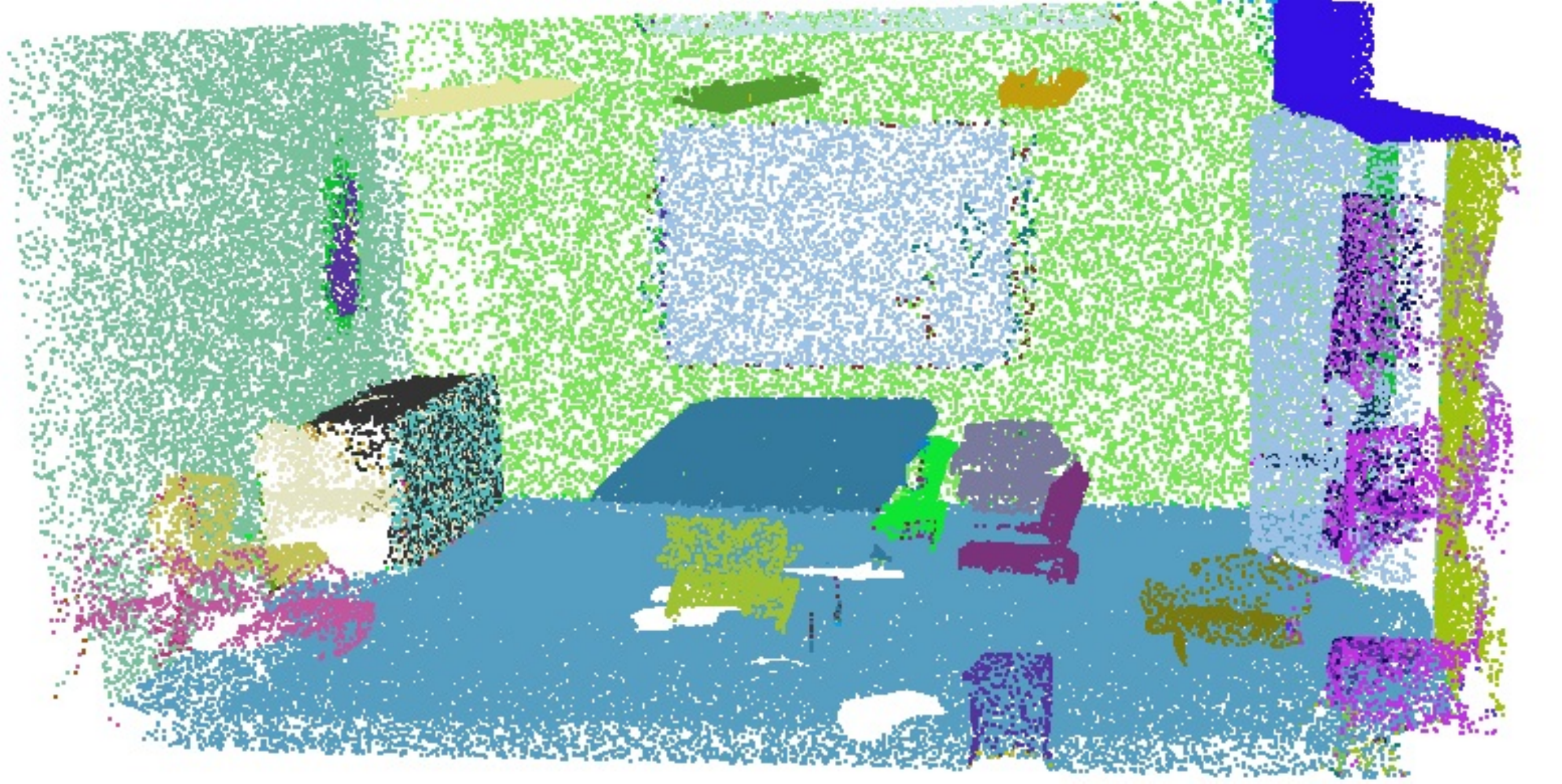}&\includegraphics[scale=0.06]{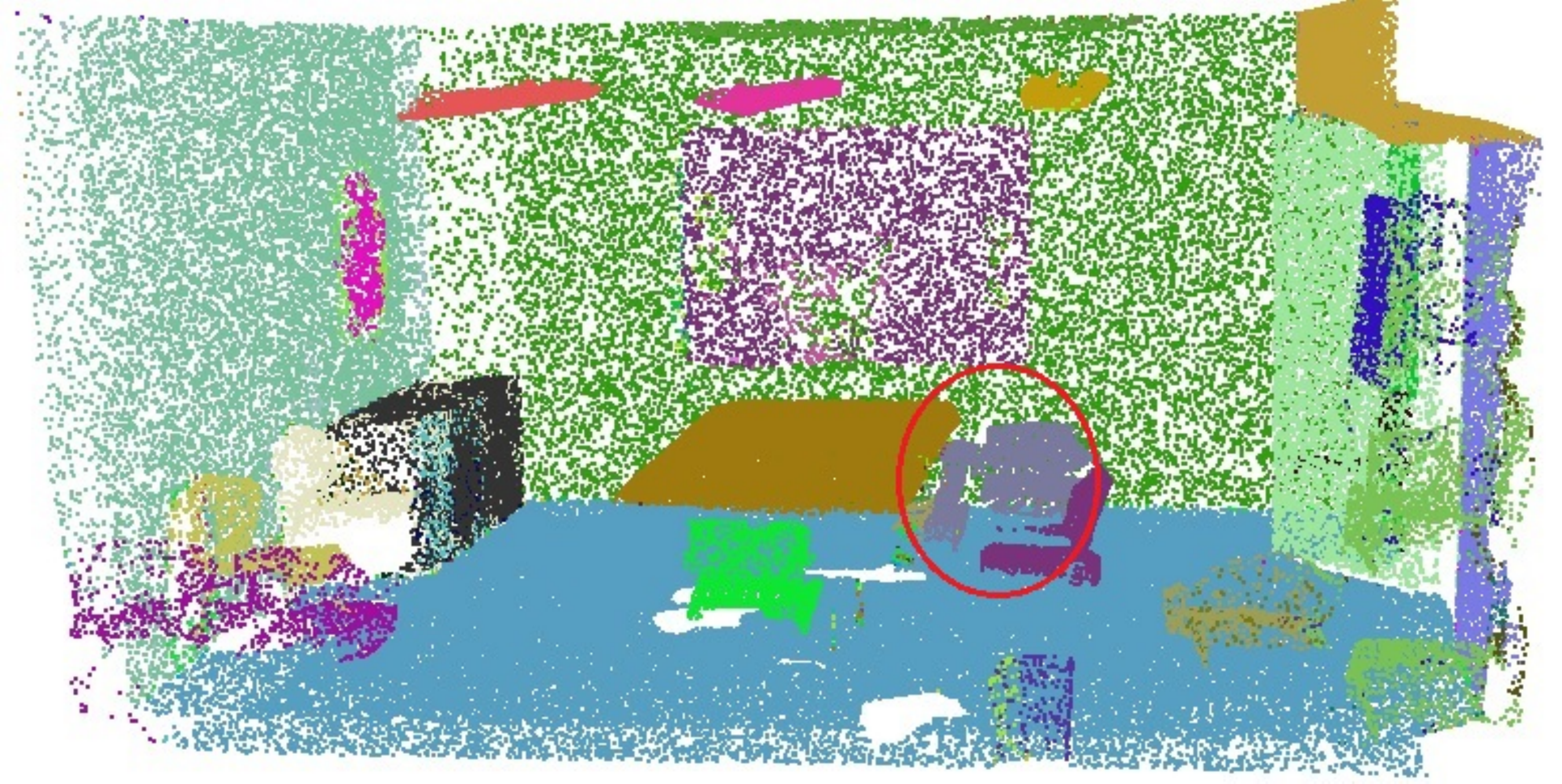}&\includegraphics[scale=0.06]       {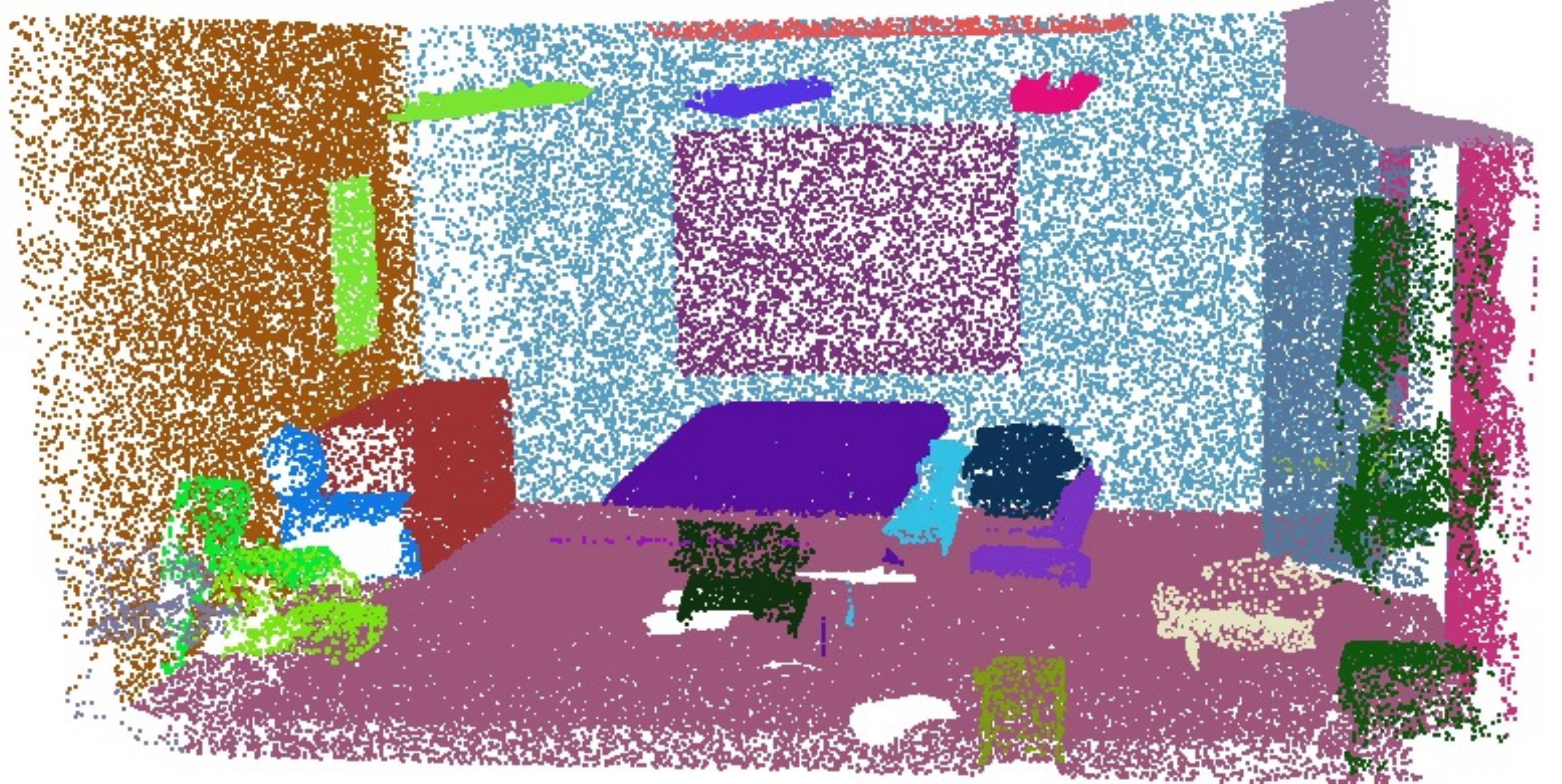}&\includegraphics[scale=0.06]{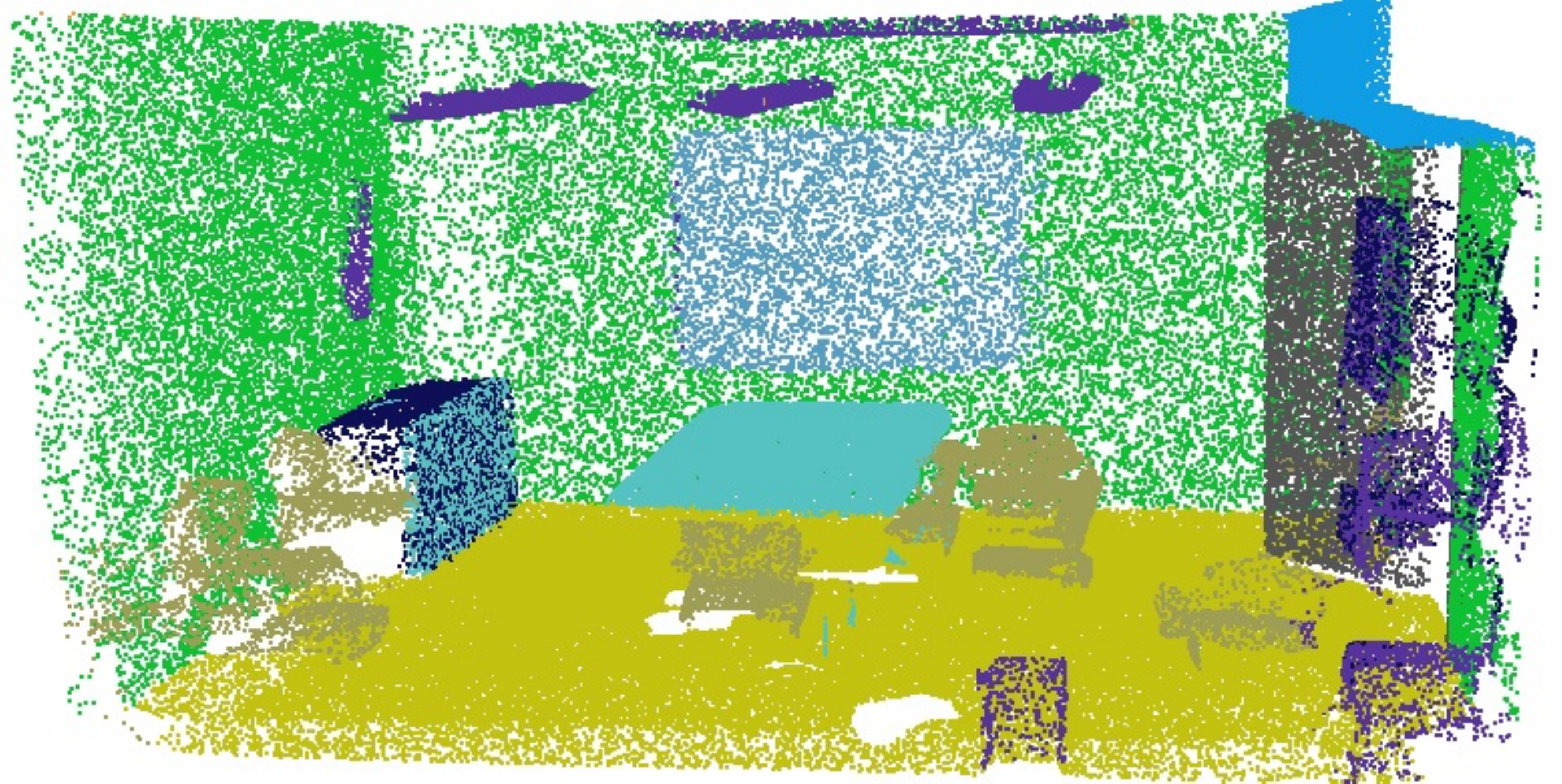}&\includegraphics[scale=0.06]{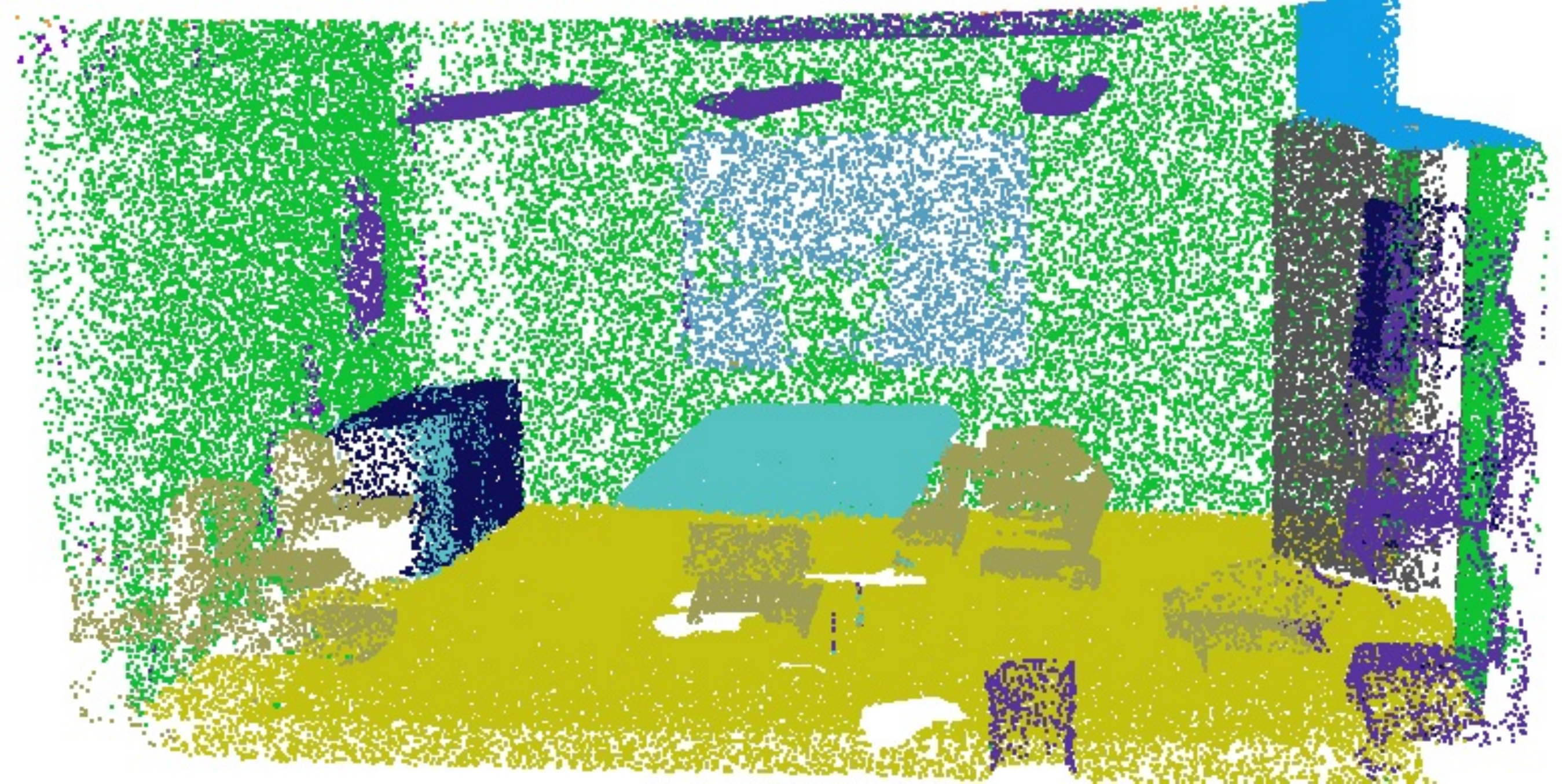}&\includegraphics[scale=0.06]{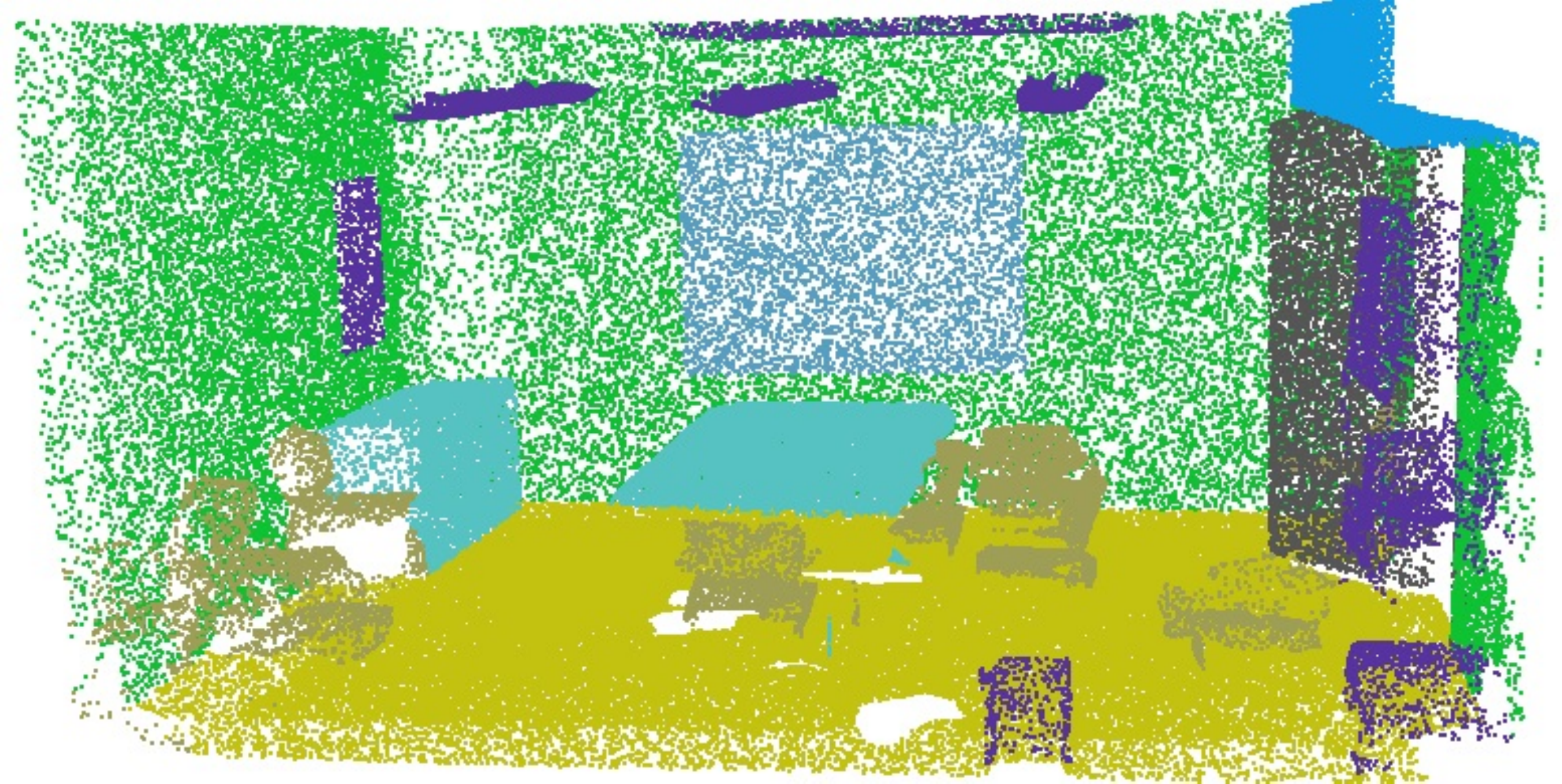}\\\\	
	\includegraphics[scale=0.08]{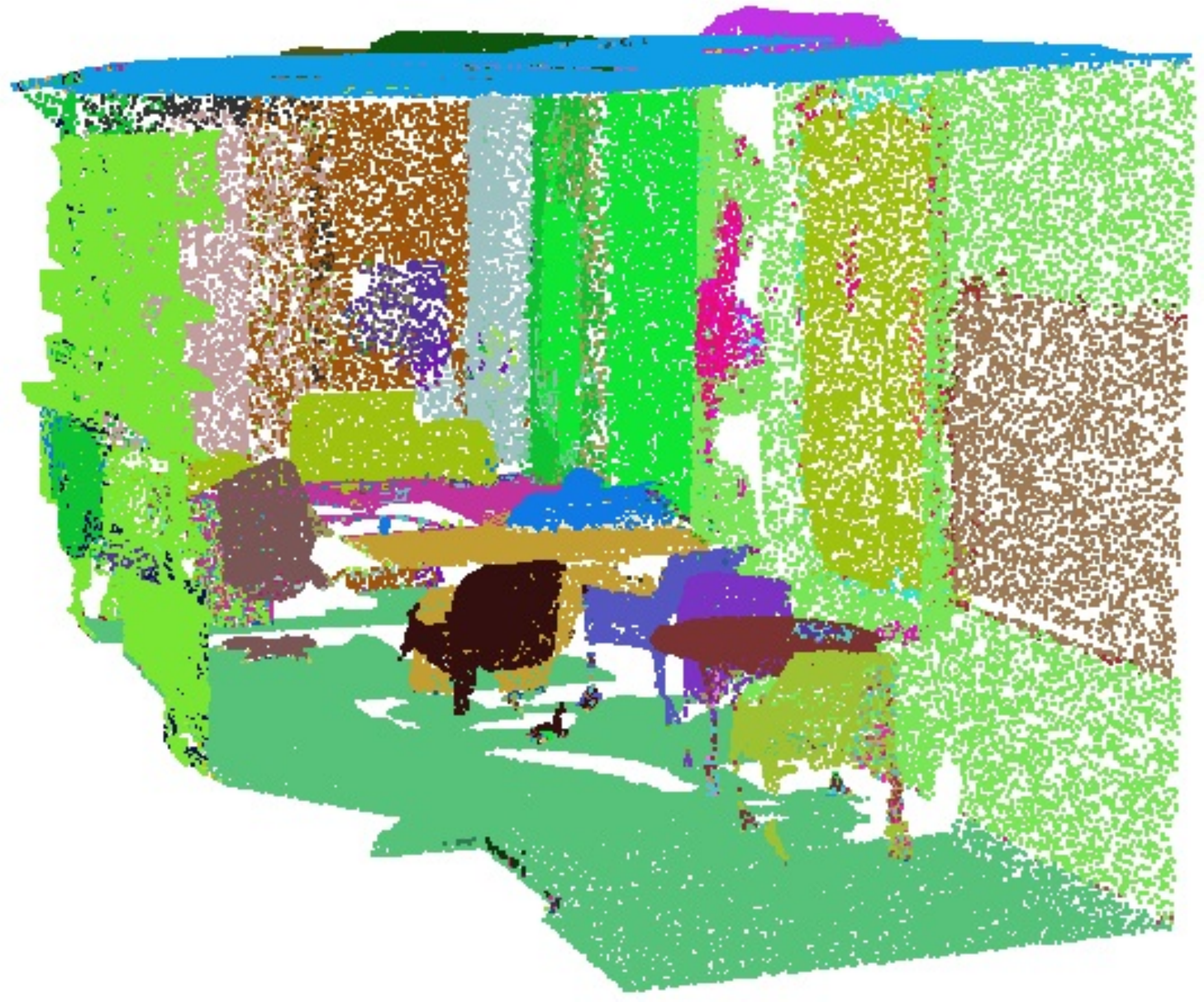}&\includegraphics[scale=0.08]{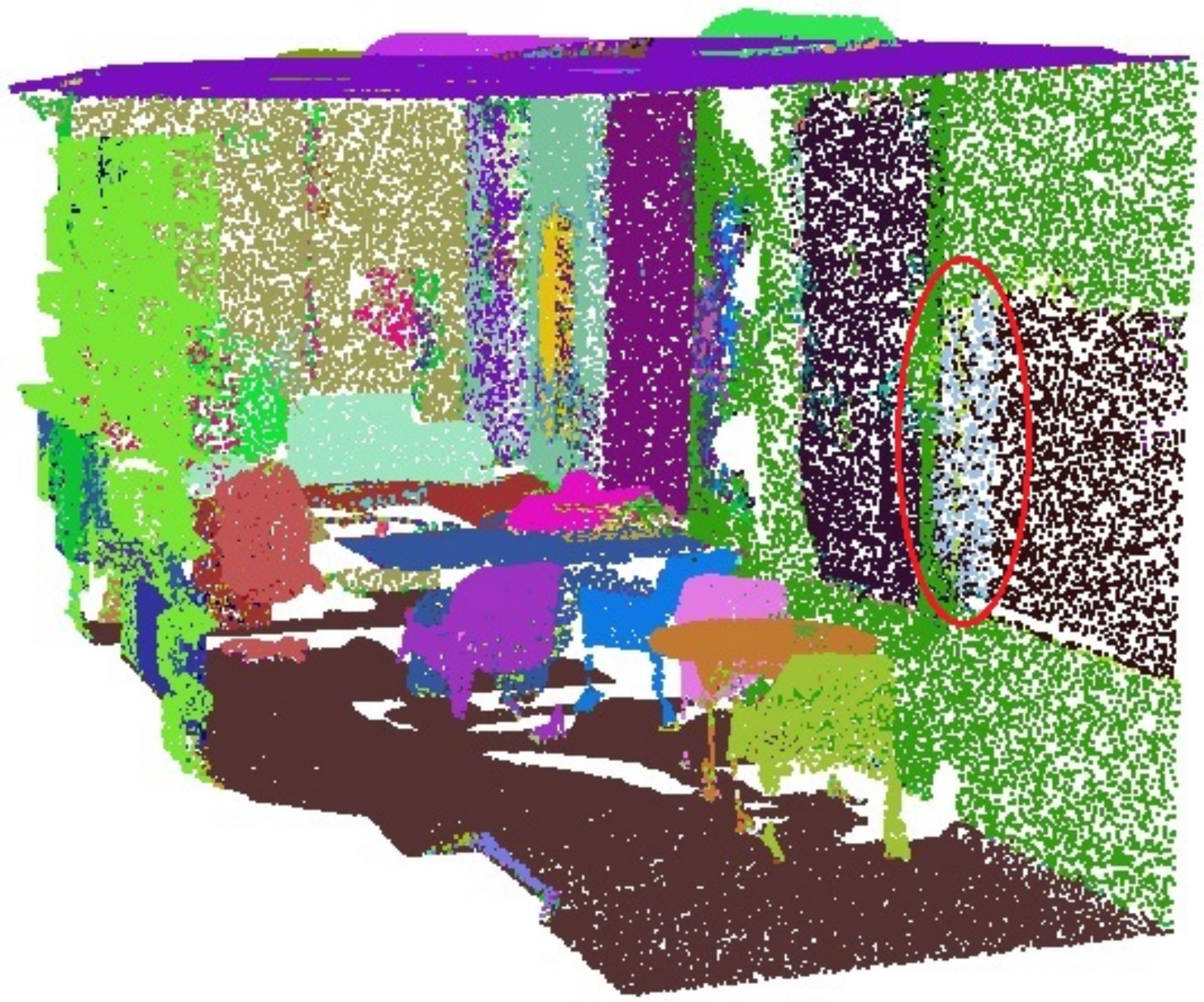}&\includegraphics[scale=0.08]       {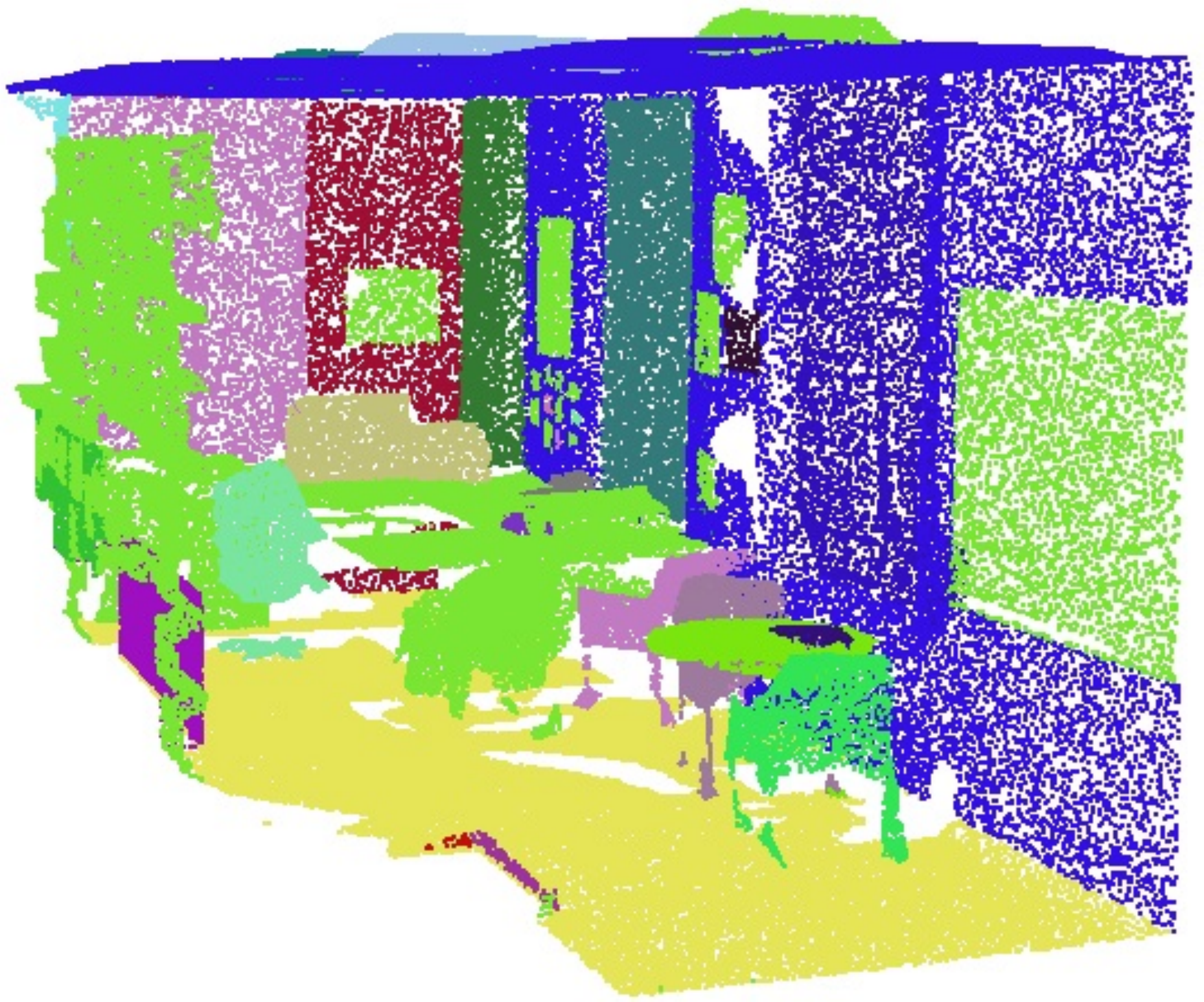}&\includegraphics[scale=0.08]{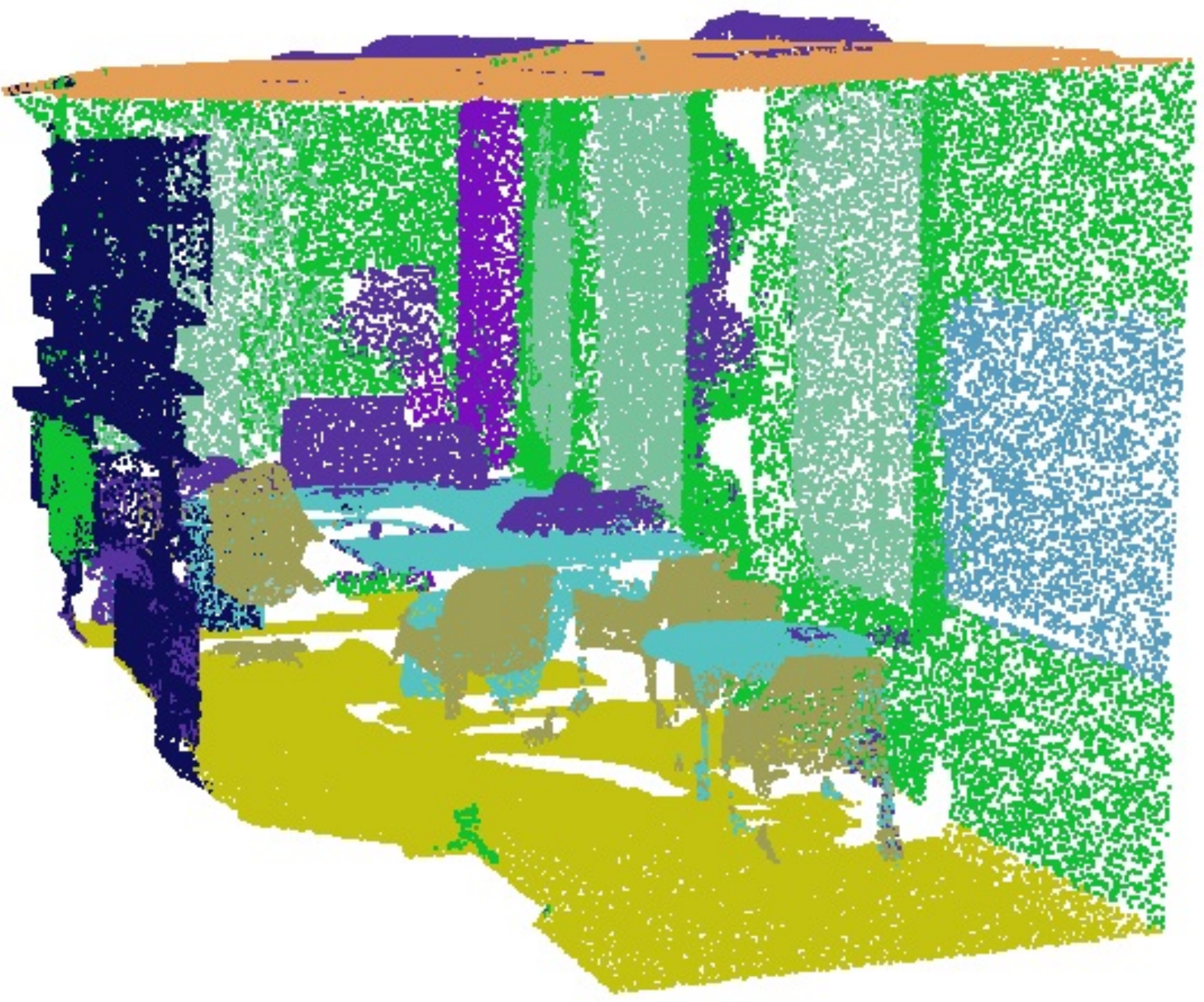}&\includegraphics[scale=0.08]{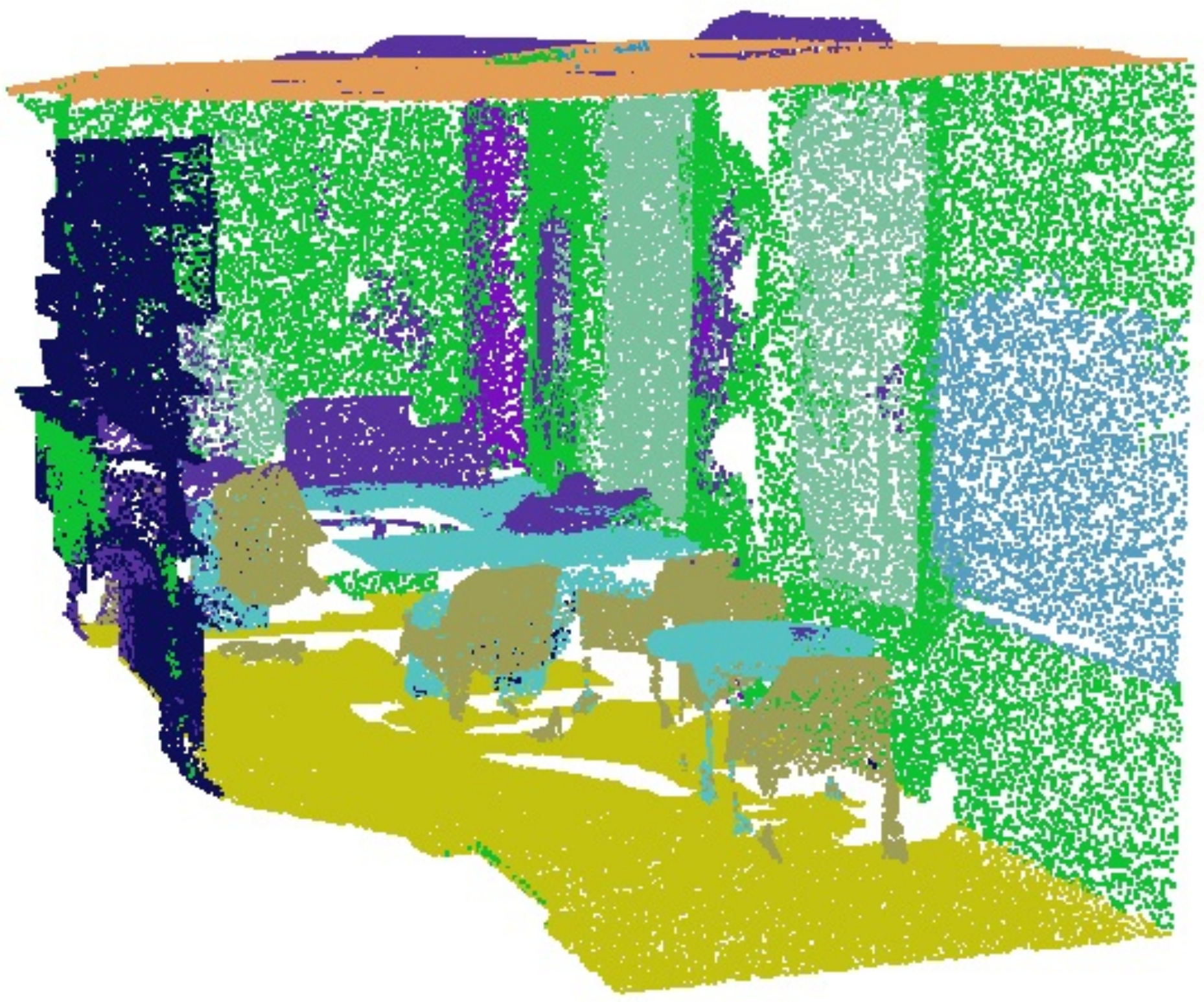}&\includegraphics[scale=0.08]{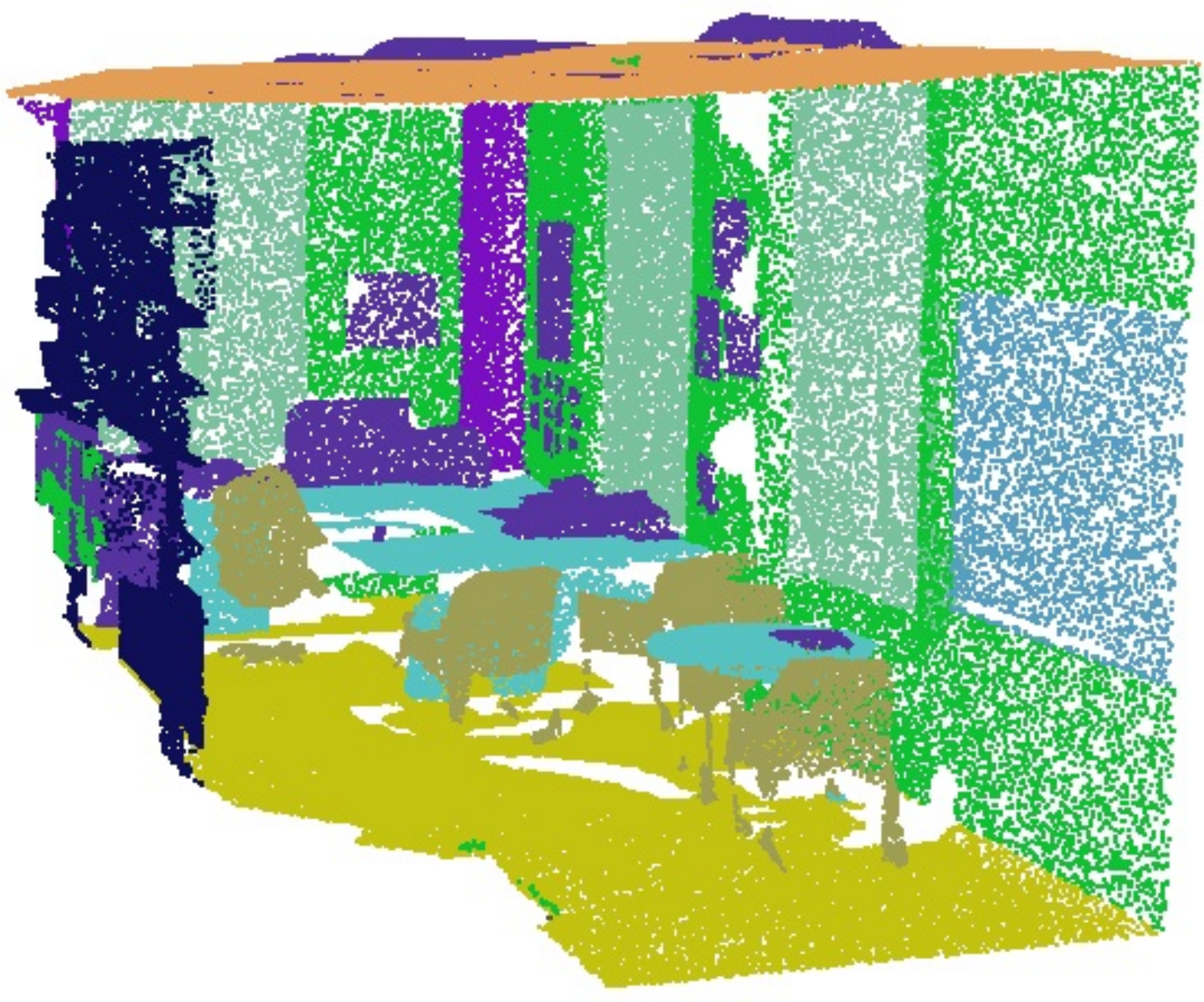}\\\\
	Ours&ASIS&GT&Ours&ASIS&GT  
\end{tabular}
\caption{Visual comparison of instance and semantic segmentation results on the S3DIS dataset. The first three columns are the instance segmentation results, while the last three columns show semantic segmentation results.}
\label{fig:S3DIS}
\end{figure}

We show some visual results of semantic and instance segmentation methods in Fig.~\ref{fig:S3DIS}. From results, we can see ours are more accurate and uniform compared with ASIS~\cite{wang2019associatively}, especially for instance segmentation as marked by red circles. We believe it is because of the applying of attention operations and the introduction of non-local information. The more studies of attention mechanisms are in Sec.~\ref{sec:dis}.

\subsection{PartNet Results}
In addition to object instance segmentation in indoor scenes, we further evaluate our method on part instance segmentation in objects using the PartNet dataset. This task is more fine-grained and thus requires more perception ability to understand the similarity between points.

The semantic and instance segmentation scores are listed in Tab.~\ref{tab:PartNet}. We can see that the performance has a significant drop compared with the previous one. This is because the dataset contains many kinds of small semantic parts, which are difficult to perceive and predict, causing low semantic mIoU and instance mCov but relative high semantic oAcc.
For this kind of dataset with small semantic parts, ASIS~\cite{wang2019associatively} with KNN is difficult to adapt by a fixed range control parameter. However, with the Bi-Directional Attention module, our method could compute the similarities between any of two points and achieves better results.

\begin{figure}[htbp]
\centering
    \begin{tabular}{p{2cm}<{\centering} p{2cm}<{\centering} p{2cm}<{\centering} p{2cm}<{\centering} p{2cm}<{\centering} p{2cm}<{\centering}}
        \includegraphics[scale=0.1]{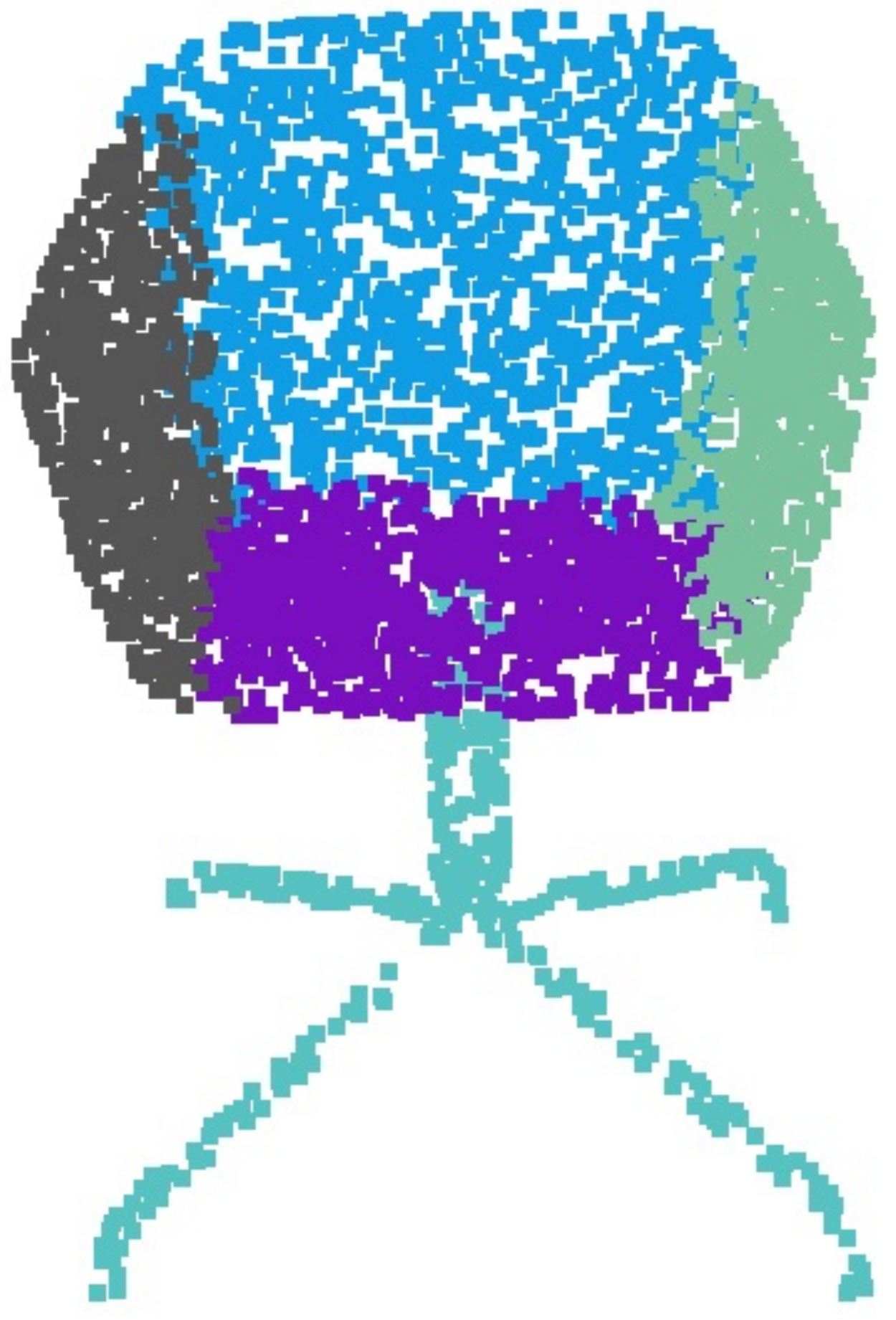}&\includegraphics[scale=0.1]{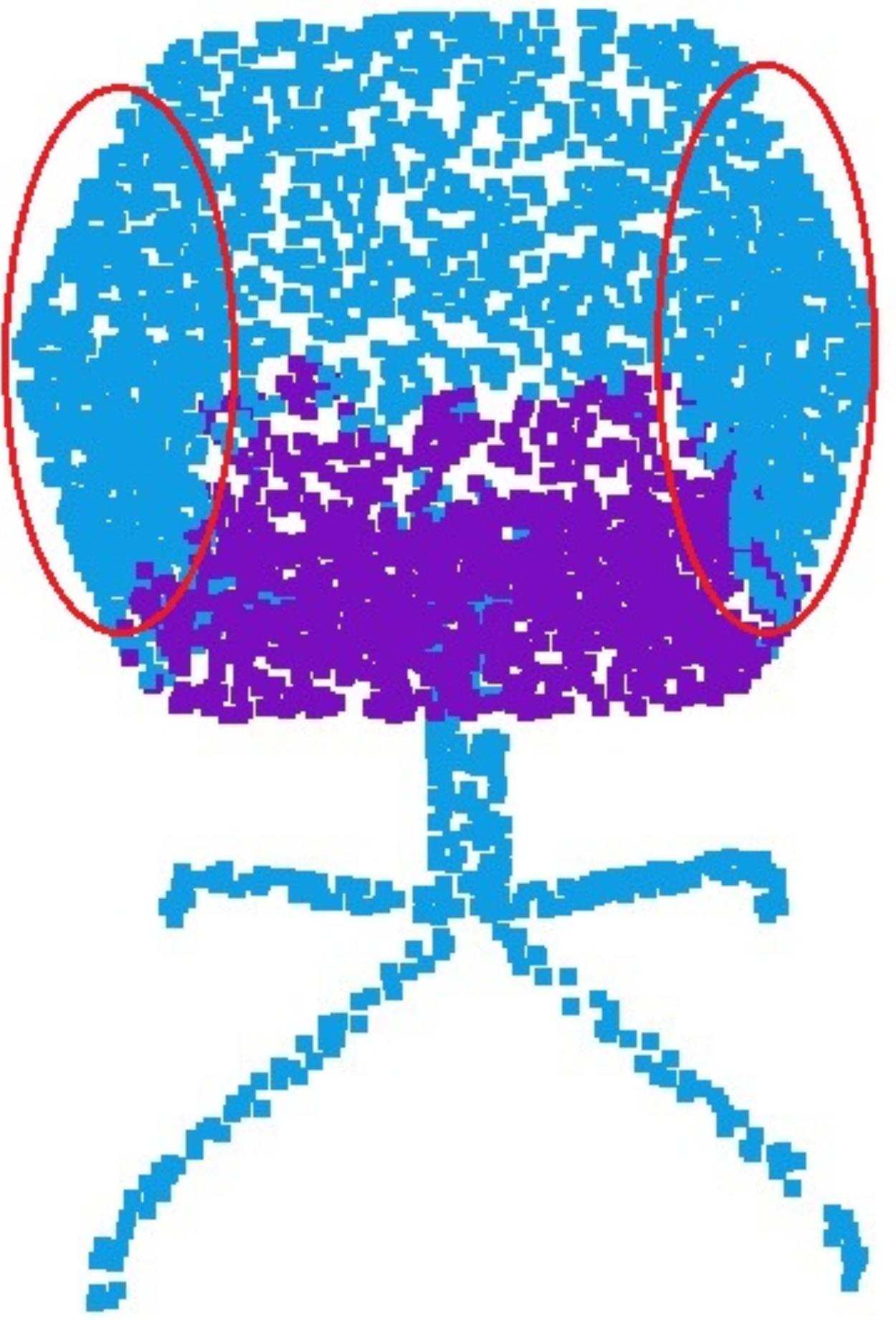}&\includegraphics[scale=0.1]       {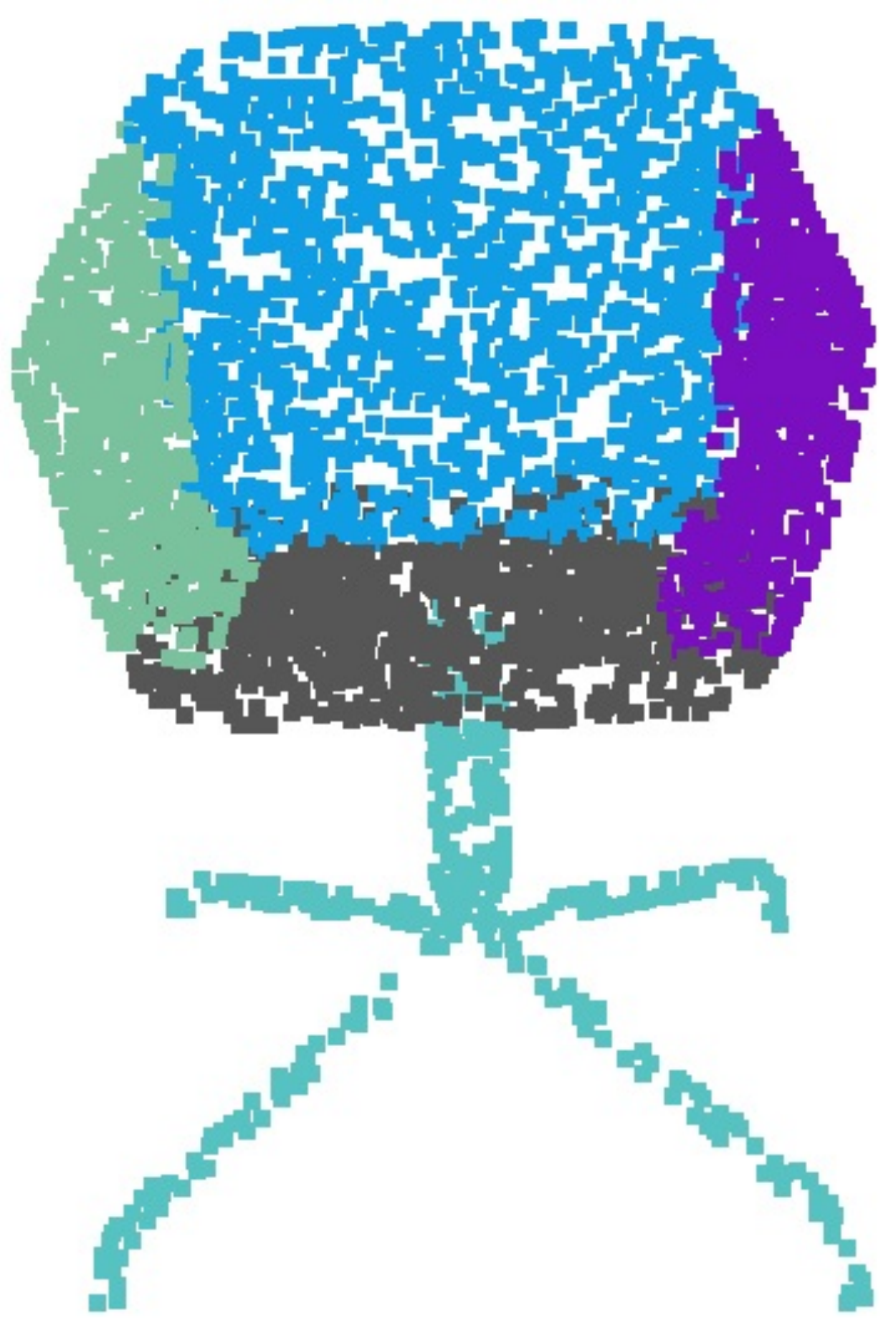}&\includegraphics[scale=0.1]{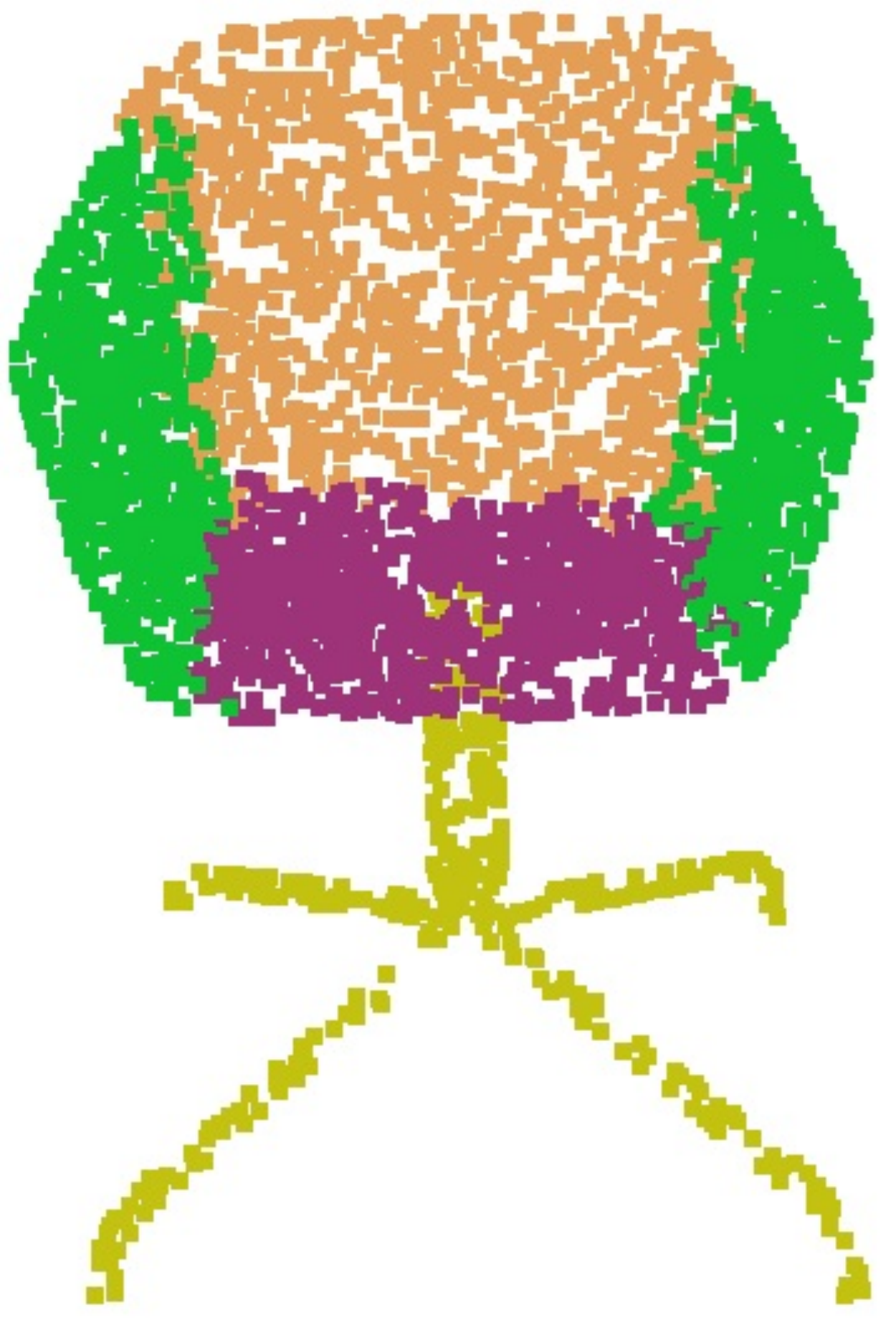}&\includegraphics[scale=0.1]{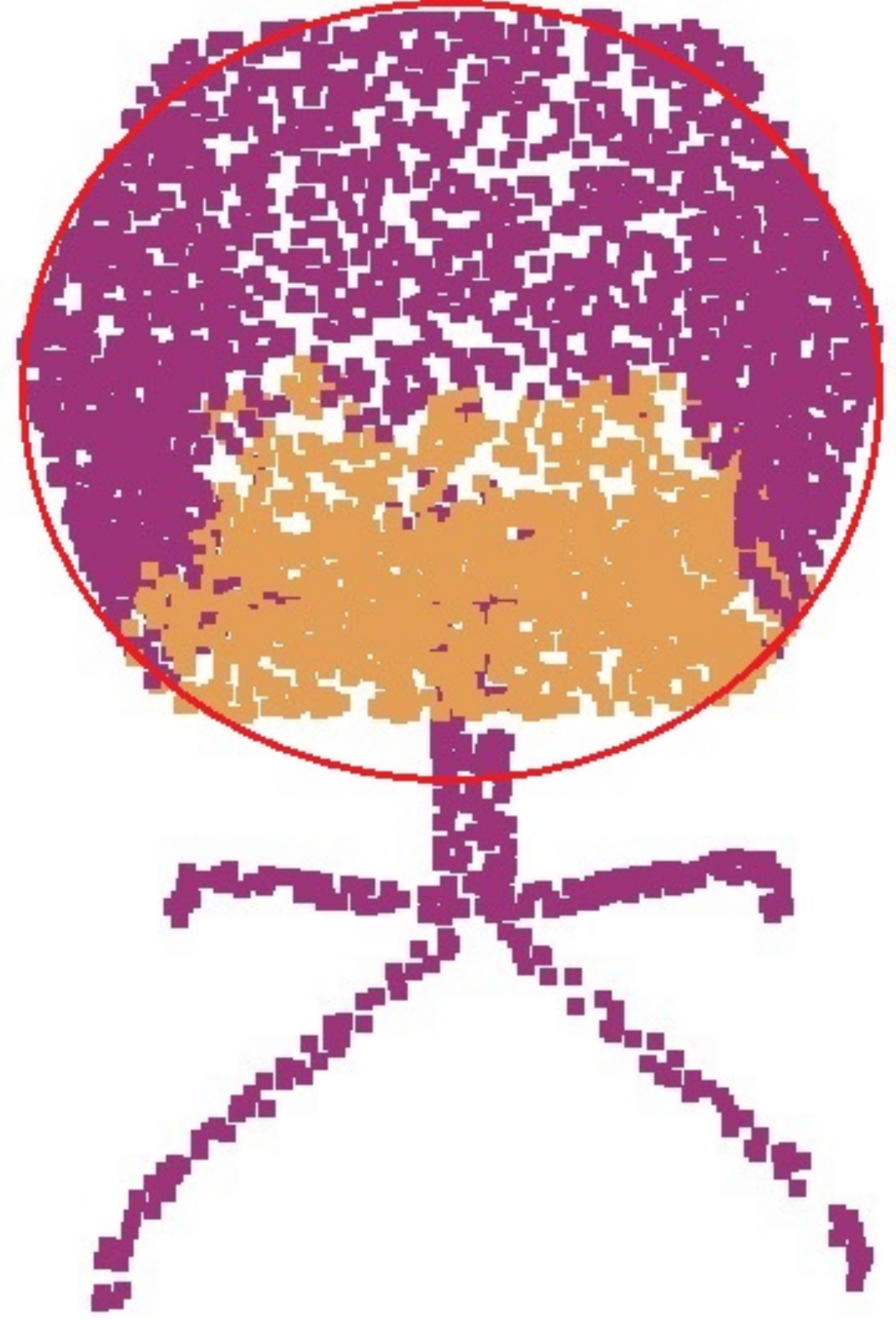}&\includegraphics[scale=0.1]{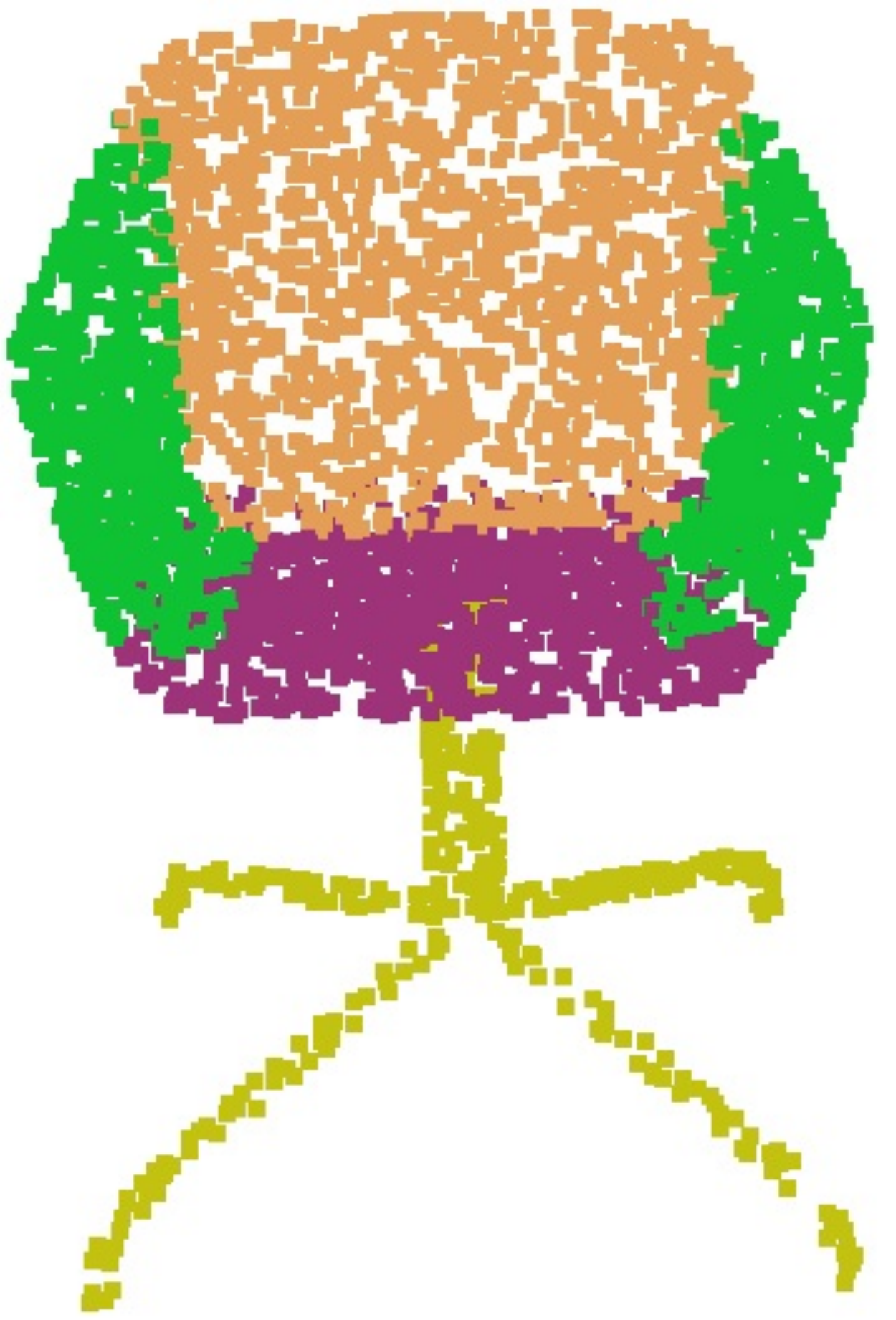}\\
    	\includegraphics[scale=0.08]{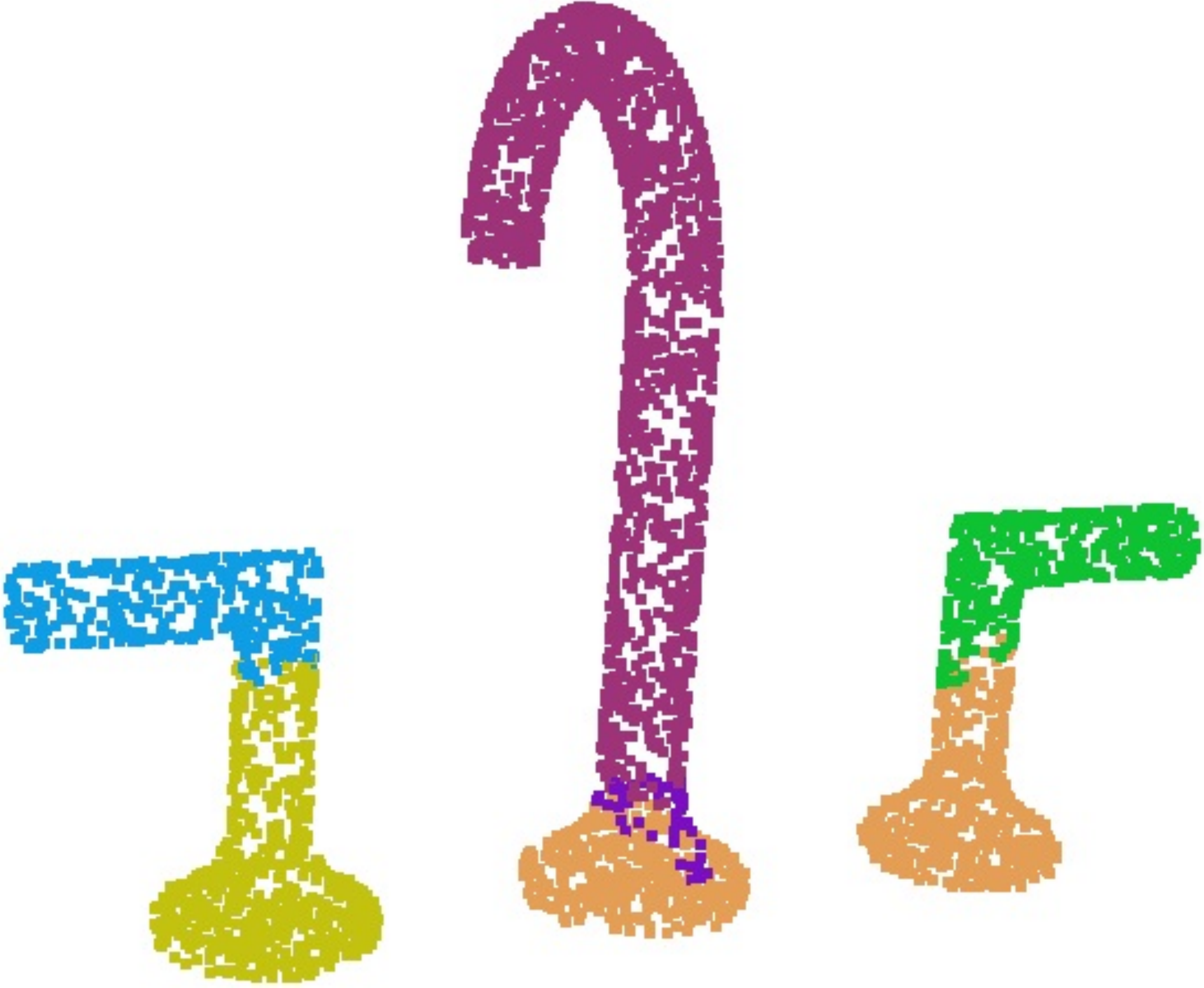}&\includegraphics[scale=0.08]{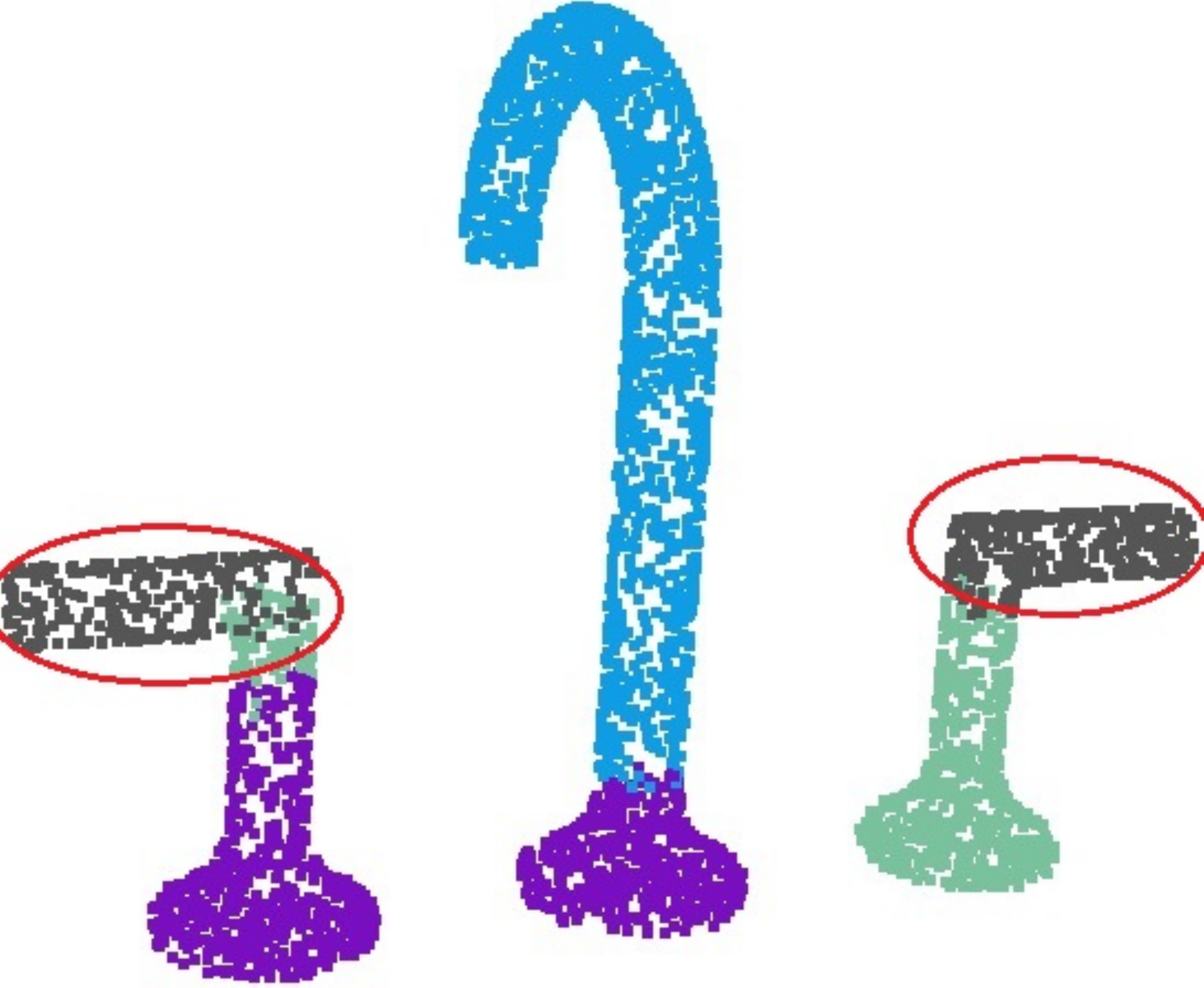}&\includegraphics[scale=0.08]       {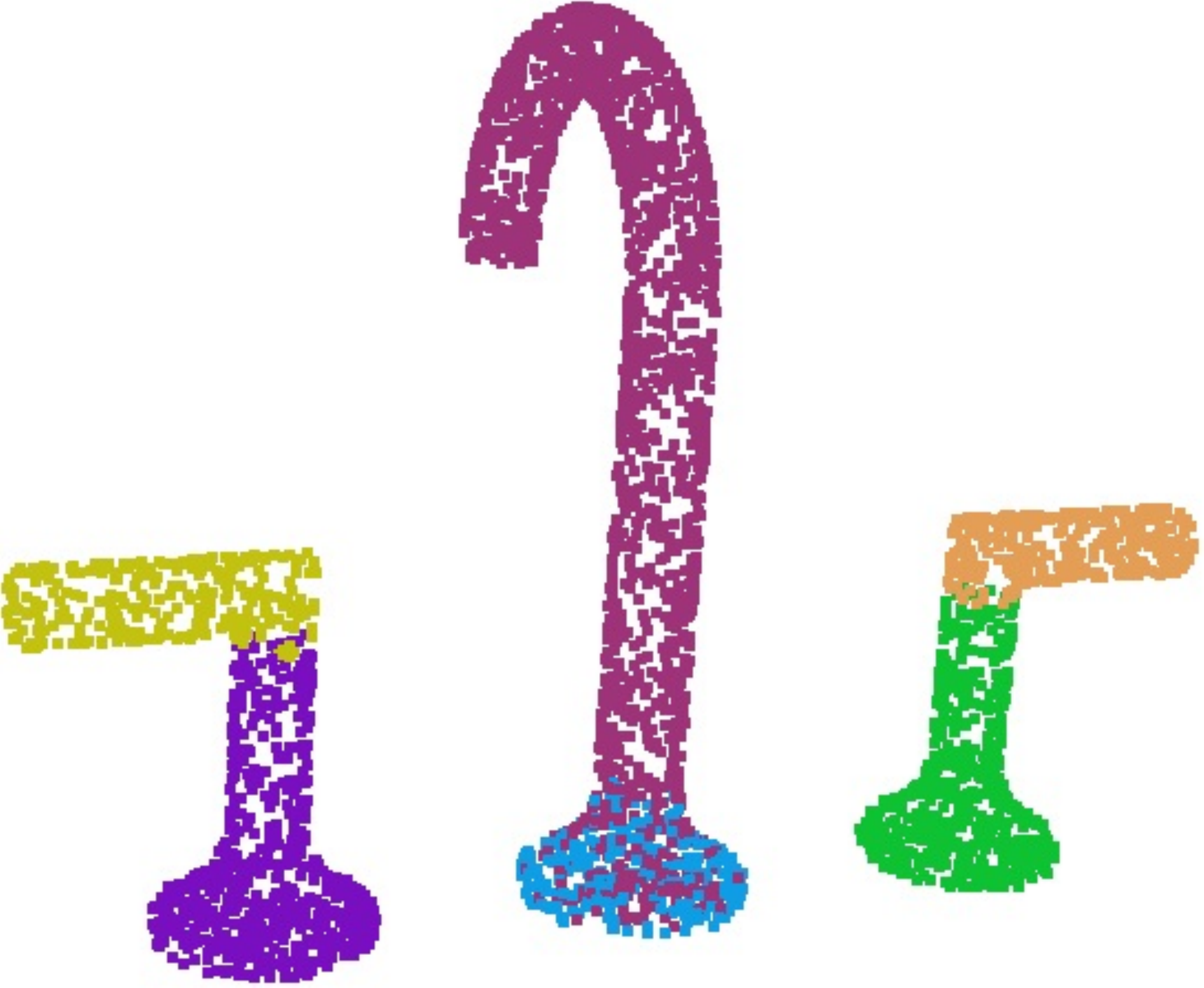}&\includegraphics[scale=0.08]{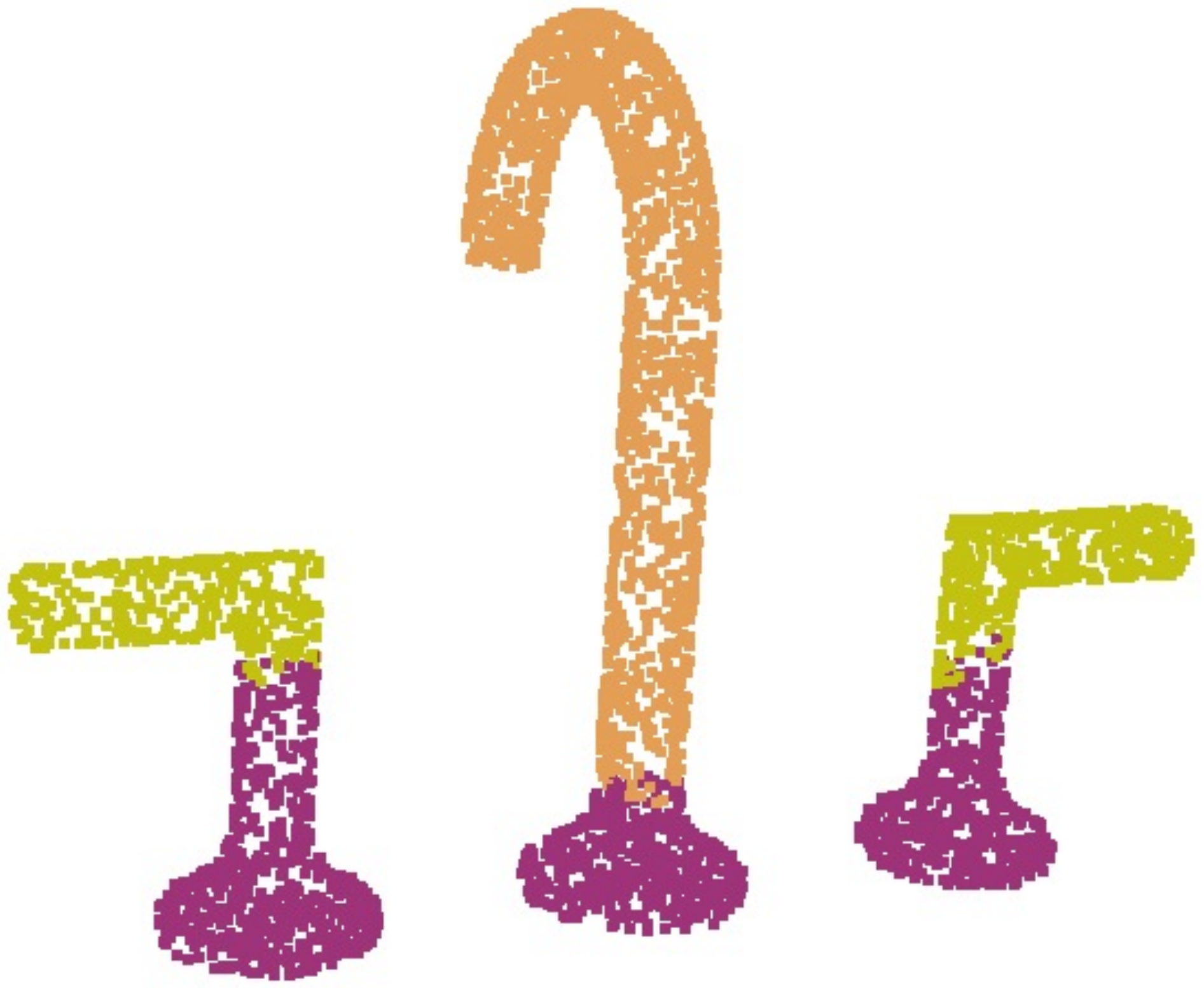}&\includegraphics[scale=0.08]{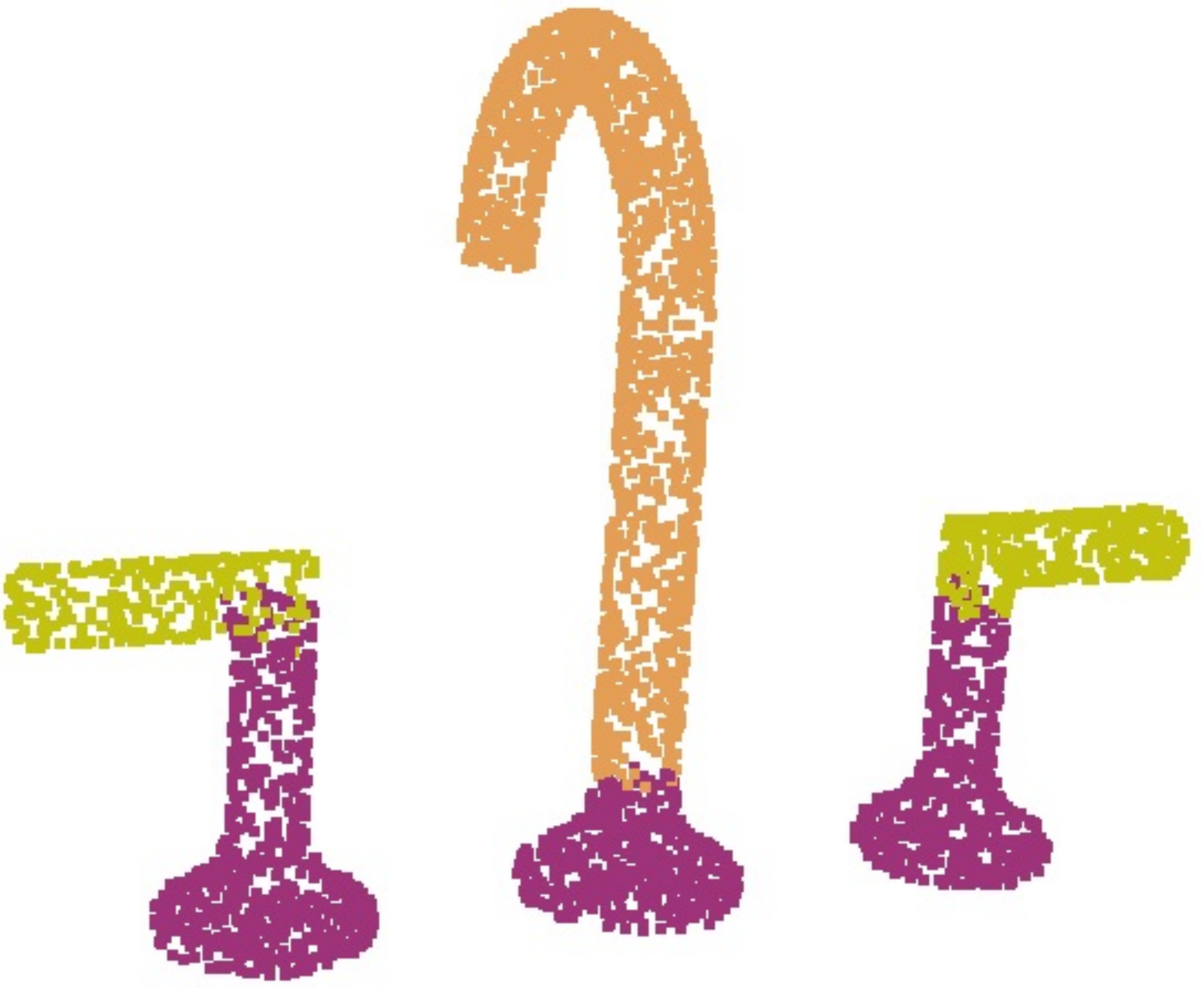}&\includegraphics[scale=0.08]{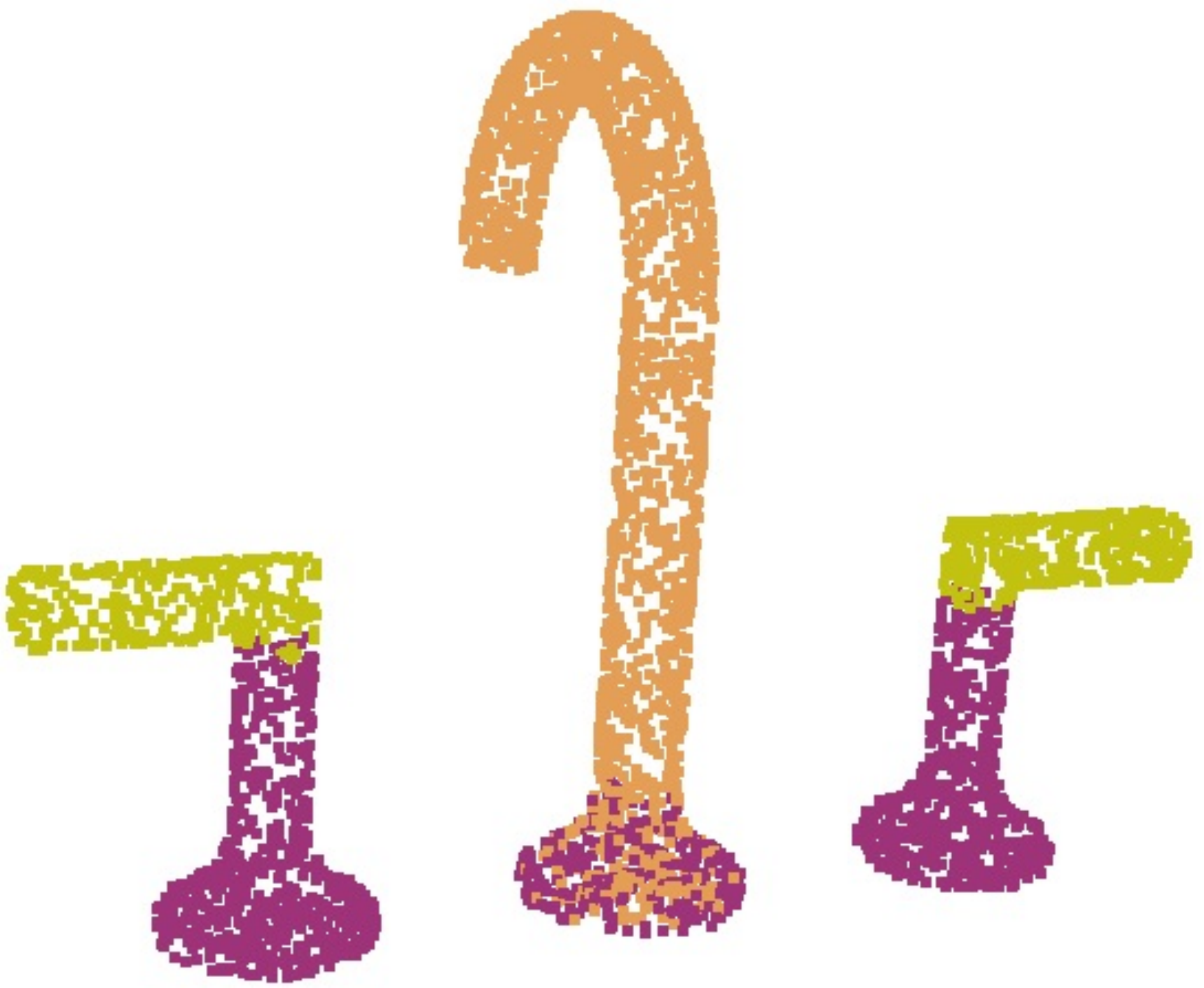}\\\\	
	    \includegraphics[scale=0.1]{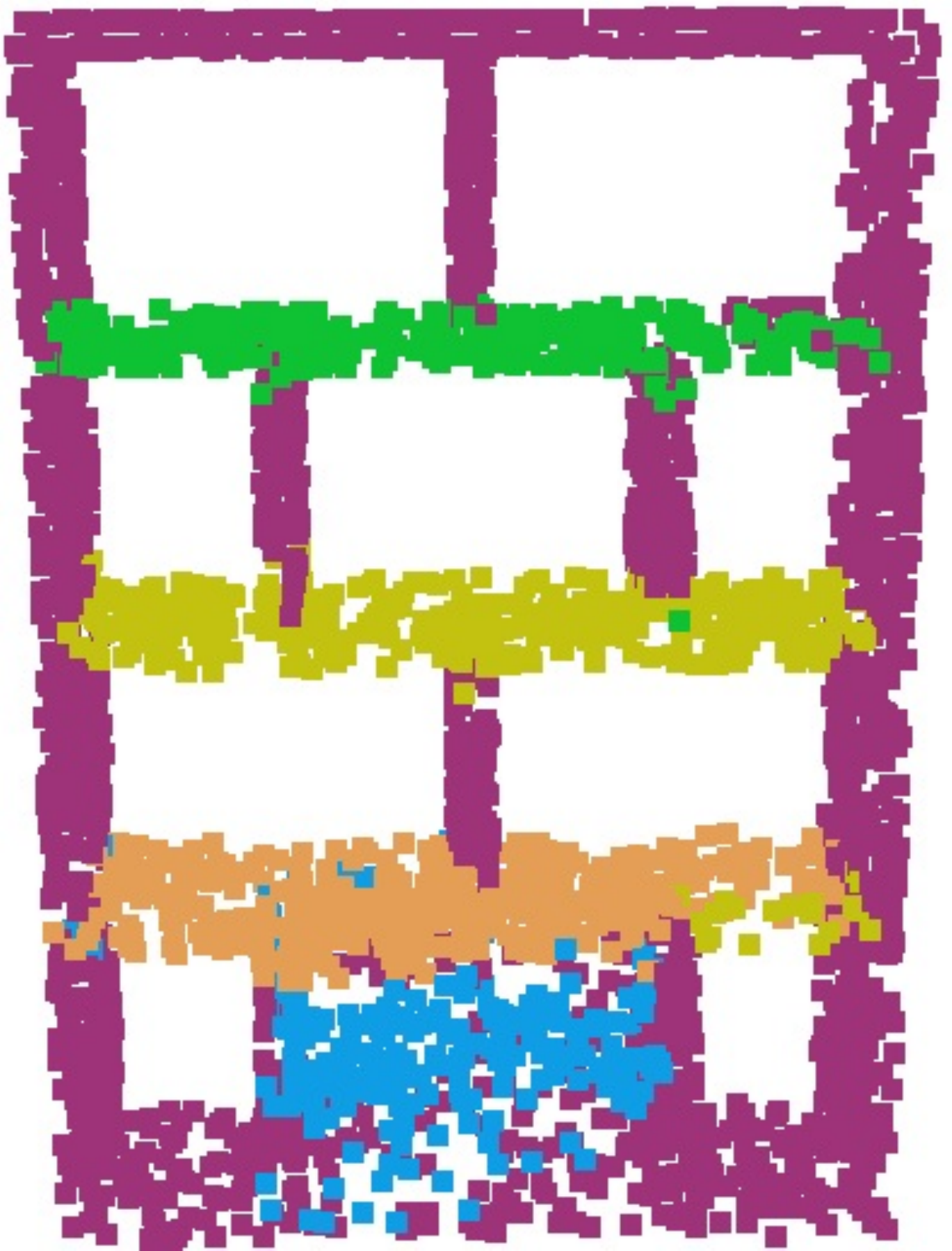}&\includegraphics[scale=0.1]{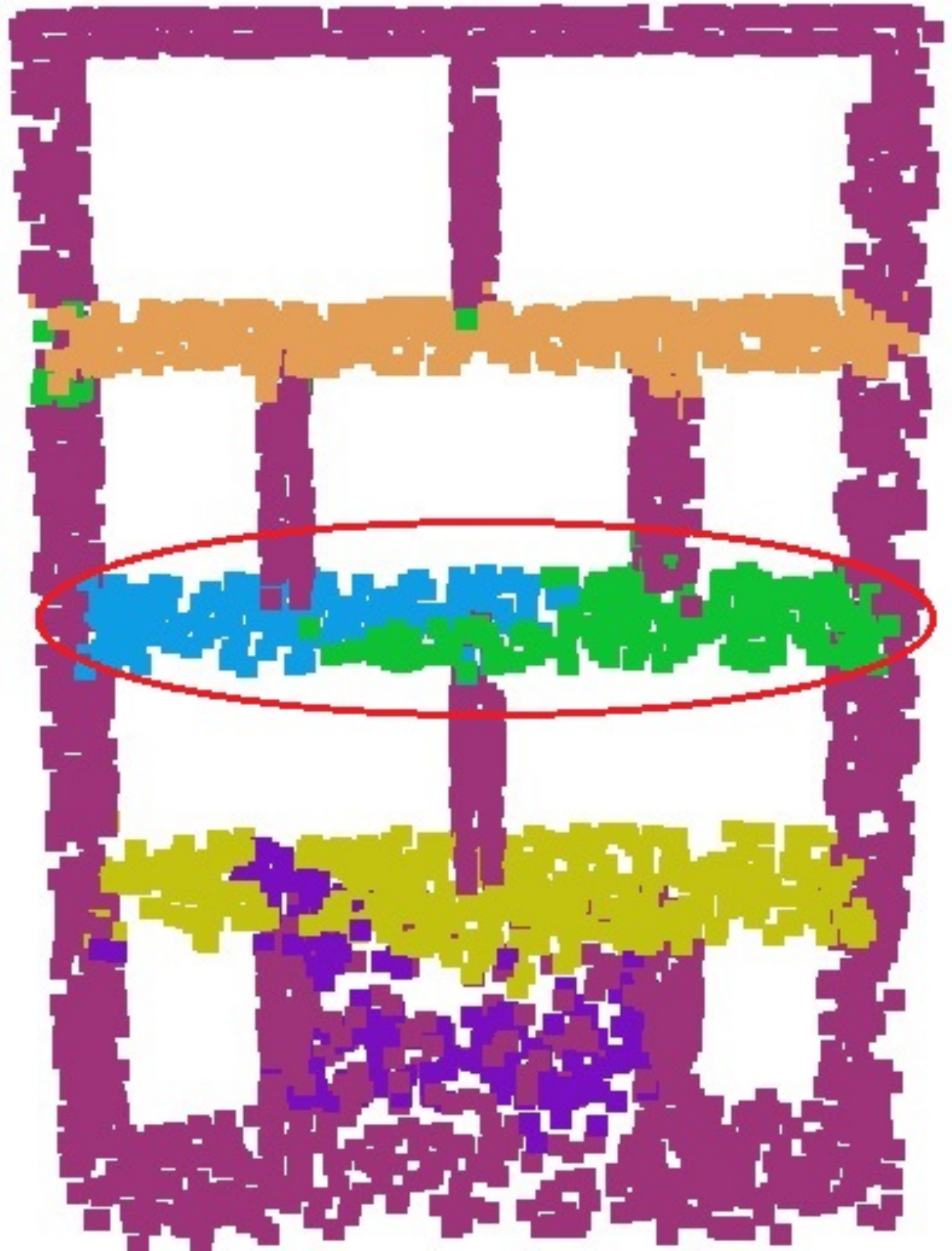}&\includegraphics[scale=0.1]       {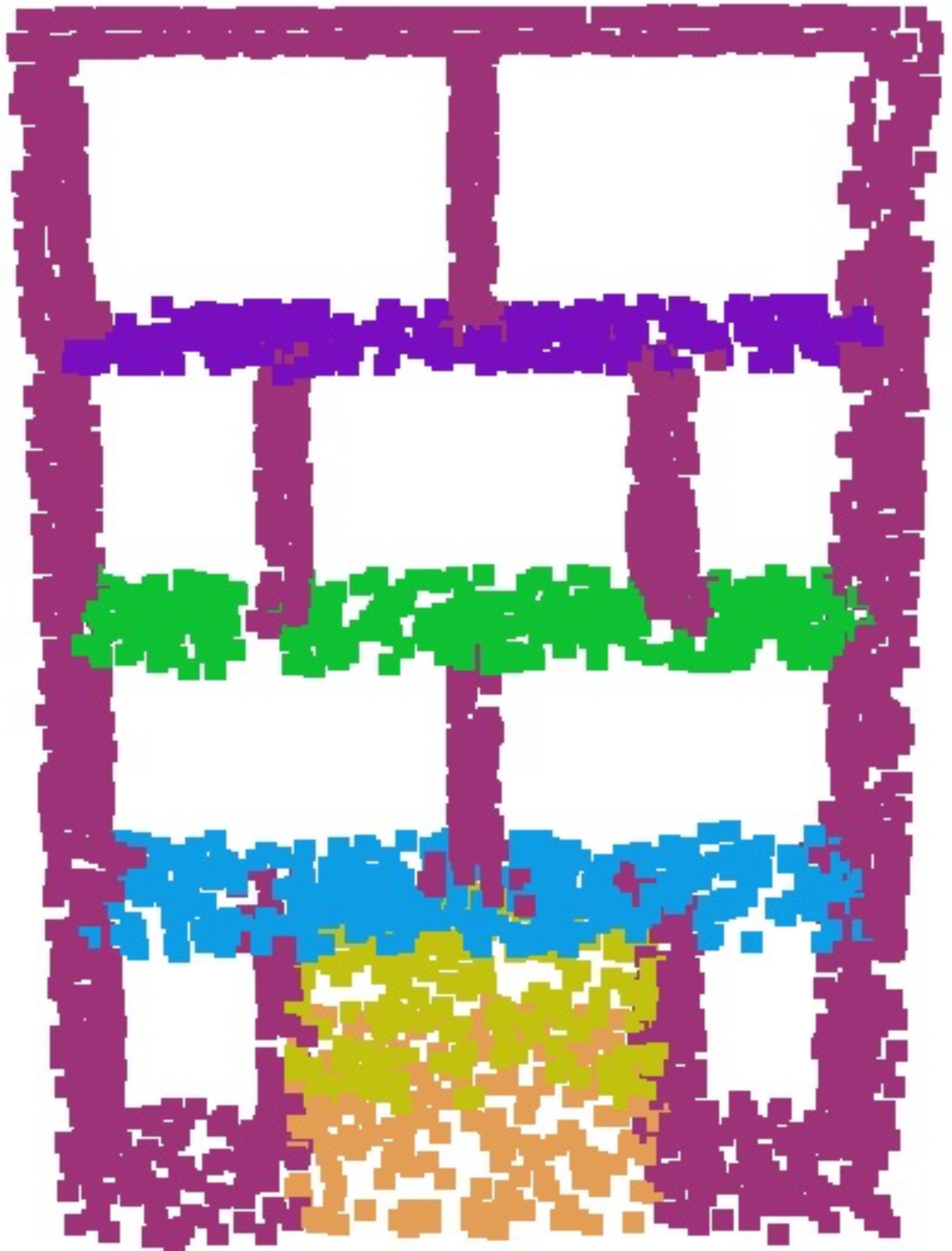}&\includegraphics[scale=0.1]{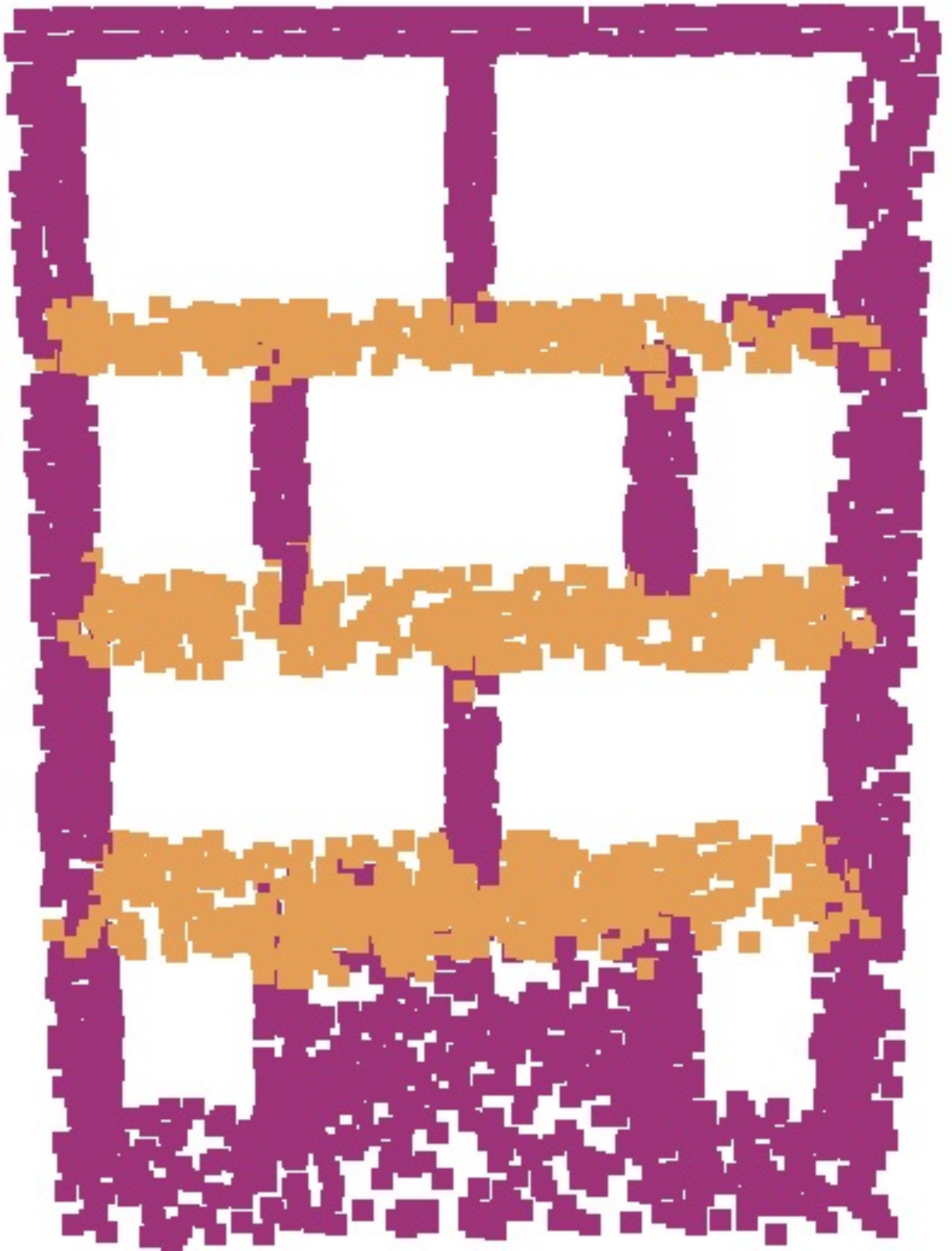}&\includegraphics[scale=0.1]{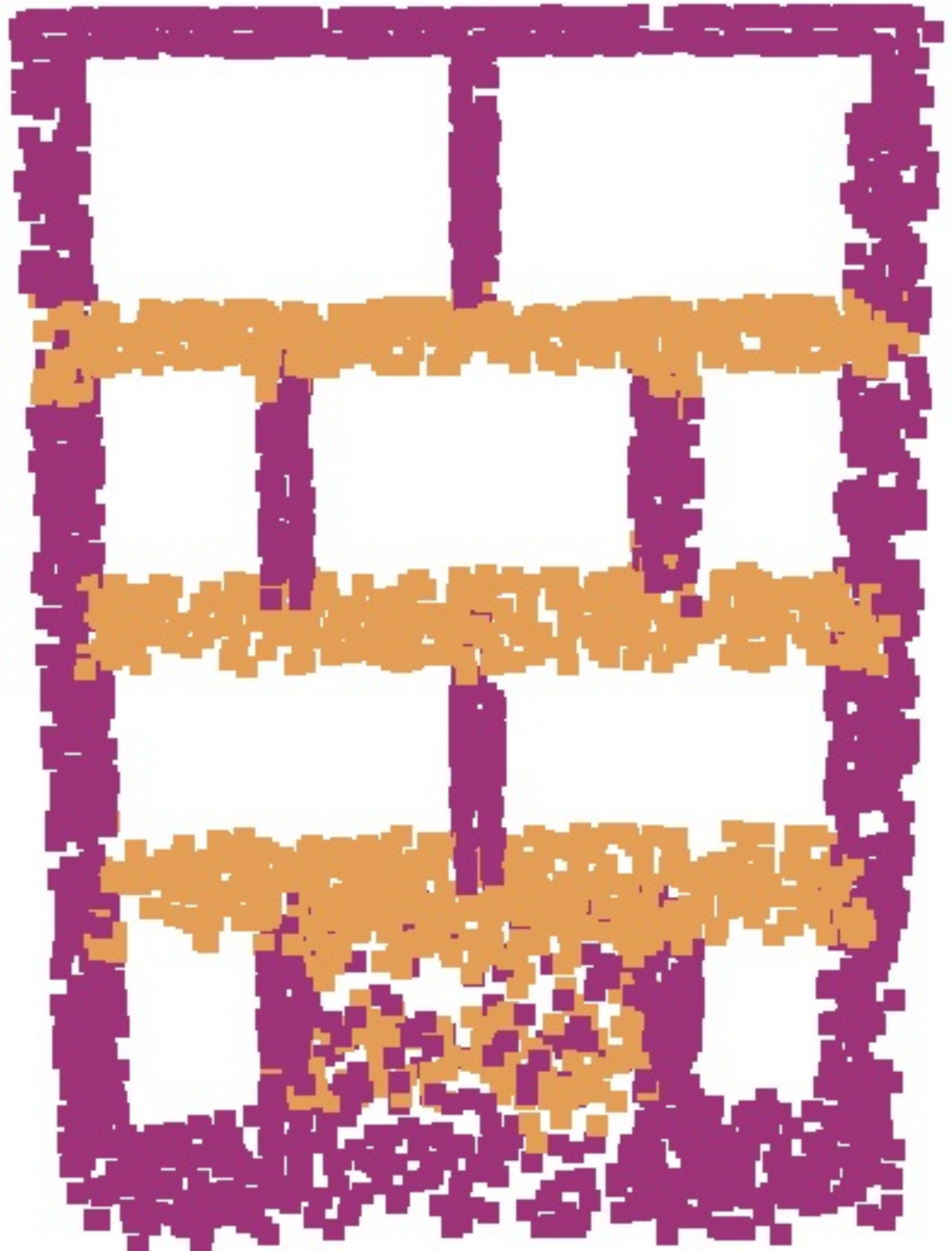}&\includegraphics[scale=0.1]{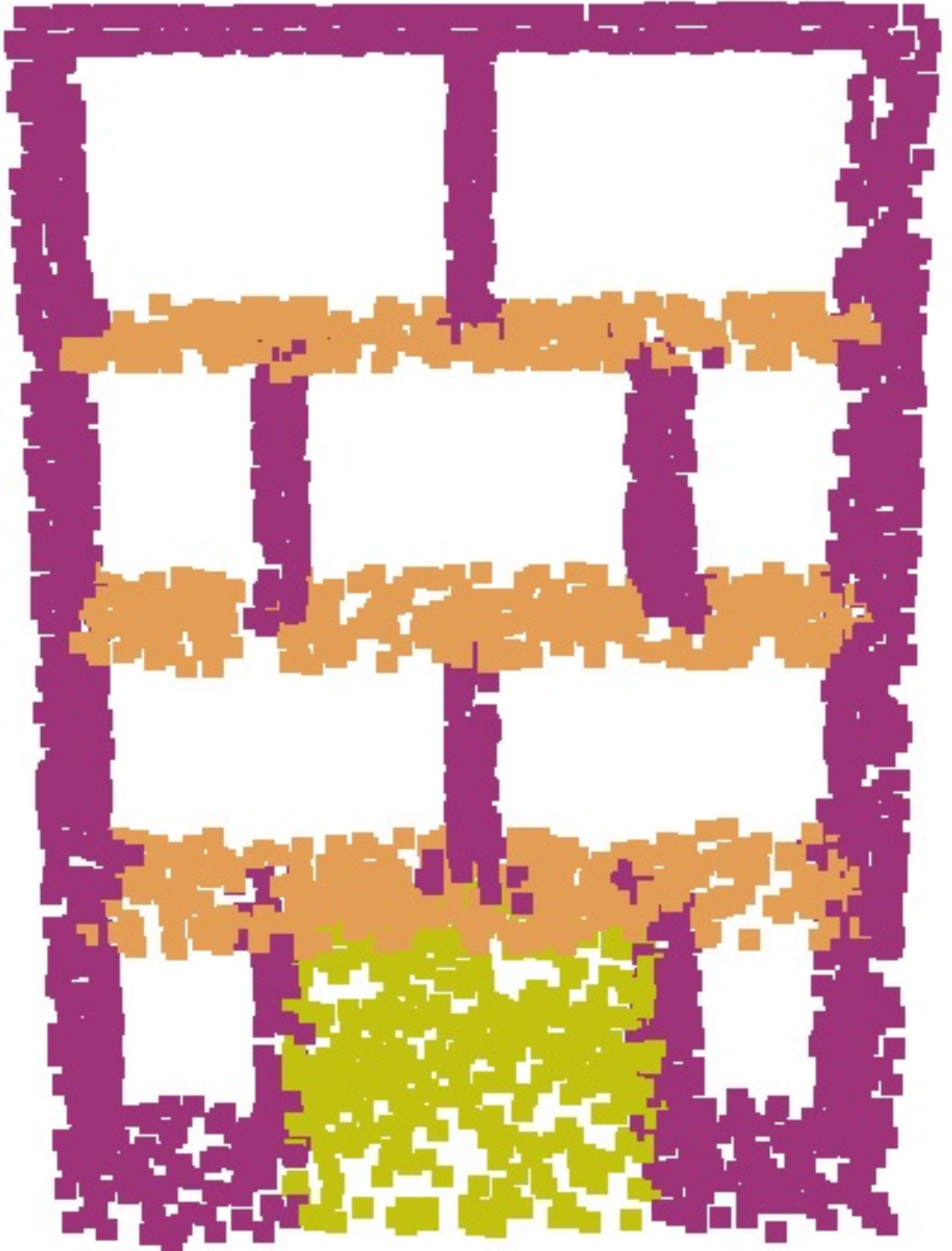}\\\\
	    Ours&ASIS&GT&Ours&ASIS&GT    
    \end{tabular}
\caption{Visual comparison of instance and semantic segmentation results on the PartNet dataset. The first three columns are the instance segmentation results, while the last three columns show semantic segmentation results.}
\label{fig:PartNet}
\end{figure}

The visual results of the PartNet dataset are shown in Fig.~\ref{fig:PartNet}. Our method demonstrates obvious advantages compared with ASIS~\cite{wang2019associatively}, and produces more accurate instance and semantic segmentation, especially for some small parts as marked by red circles.
\begin{table}[htbp]
    \centering
    \setlength{\belowcaptionskip}{5pt}
    \caption{Result on PartNet}
    \begin{tabular}{c|c|c|c|c}
        \hline\hline
        method&backbone&mCov&mIoU&oAcc\\
        \hline\hline
        PointNet++&PointNet++&42.0&43.4&78.4\\
        \hline
        ASIS&Pointnet++&39.3&40.2&76.7\\
        \hline
        Our model&Pointnet++&\textbf{42.7}&\textbf{44.9}&\textbf{80.3}\\
        \hline
        \end{tabular}
        \label{tab:PartNet}
    \vspace{-5mm}
\end{table}

\section{Discussion}\label{sec:dis}
In this section, we intend to show more evidence to justify the design and the mechanism of the proposed Bi-Directional Attention module. 
\subsection{Ablation study}
As mentioned in Sec.~\ref{sec:networks}, there are three kinds of sequences to conduct STOI and ITOS in our Bi-Directional Attention module, and we gave an assumption to decide our design. Here, we will verify our choice and further prove the necessity to have both STOI and ITOS.

In Tab.~\ref{tab:ablation}, we give five rows of results for instance and semantic segmentation with different combinations and order of STOI and ITOS. The experiments are conducted on Area 5 of S3DIS~\cite{armeni20163d}. We can see, by introducing STOI, the instance segmentation gets boosted. With ITOS, both instance and semantic segmentation demonstrate certain improvement, which suggests fusing instance features for semantic segmentation in our way is very effective. Moreover, considering the potential task conflict when using simple element-wise feature aggregation strategies such as adding and concatenating, the improvement is more significant. Finally, with both STOI and ITOS, and STOI first, we achieve the best results. But, with an inverse order that ITOS first, the performance shows a large drop, even worse than results without STOI and ITOS. This phenomenon verified the importance of order to conduct STOI and ITOS and is worth to be studied further in the future.

Further, we test performance when $X=Y$ in Eq.~\ref{eq:eq1} where our Bi-Directional Attention module is degraded to two independent self-attention operations~\cite{wang2018non}. The result is listed in the last row of Tab.~\ref{tab:ablation}. Obviously, without feature fusing, self-attention is not comparable to our method.





\begin{table}
    \centering
    \setlength{\belowcaptionskip}{5pt}
    \caption{Results of all ablation experiments on Area 5 of S3DIS.}
    \begin{tabular}{c|c|c|c|c|c|c|c|c}
         \hline \hline
         \multicolumn{2}{c}{Ablation}&\multicolumn{4}{|c|}{Instance segmentation}&\multicolumn{3}{c}{Semantic segmentation}\\
         \hline \hline
         STOI&ITOS&mCov&mWCov&mPrec&mRec&mAcc&mIoU&oAcc\\
         \hline
         $\times$&$\times$&46.0&49.1&54.2&43.3&62.1&53.9&87.3\\
         \hline
         \checkmark&$\times$&47.1&50.1&55.3&43.6&61.2&53.4&87.0\\
         \hline
         $\times$&\checkmark&47.4&50.3&54.0&43.4&62.0&54.7&87.8\\
         \hline
         \checkmark&\checkmark&\textbf{49.0}&\textbf{52.1}&\textbf{56.7}&\textbf{45.9}&\textbf{62.5}&\textbf{55.2}&87.7\\
         \hline
         \multicolumn{2}{c|}{Inverse order}&46.3&49.4&53.5&41.5&\textbf{62.5}&55.1&\textbf{87.9}\\
         \hline\hline
         \multicolumn{2}{c|}{Self-attention}&45.4&48.6&53.3&43.6&\textbf{62.5}&55.1&\textbf{87.9}\\
         \hline
    \end{tabular}
    \label{tab:ablation}
\end{table}

\subsection{Mechanism Study}
Here, we visualize the learned instance and semantic similarity matrices $P$ defined in Eq.~\ref{eq:eq1} to study and verify their mechanism.
The similarity matrix is the key functional unit, which builds the pair-wise similarities and uses to weighted-sum non-local information.
A good instance similarity matrix should accurately reflect the similarity relationship between all of the points, so $P$ are of size $N\times{N}$. When the instance/semantic similarity matrix trained well, it will help generate uniform and robust semantic/instance features. Besides, good instance and semantic similarity matrices will also benefit the back-propagation process, as stated in Sec.~\ref{sec:grad}. 

\begin{figure}[htbp]
\centering
    \begin{tabular}{cccc}
        \includegraphics[scale=0.141]{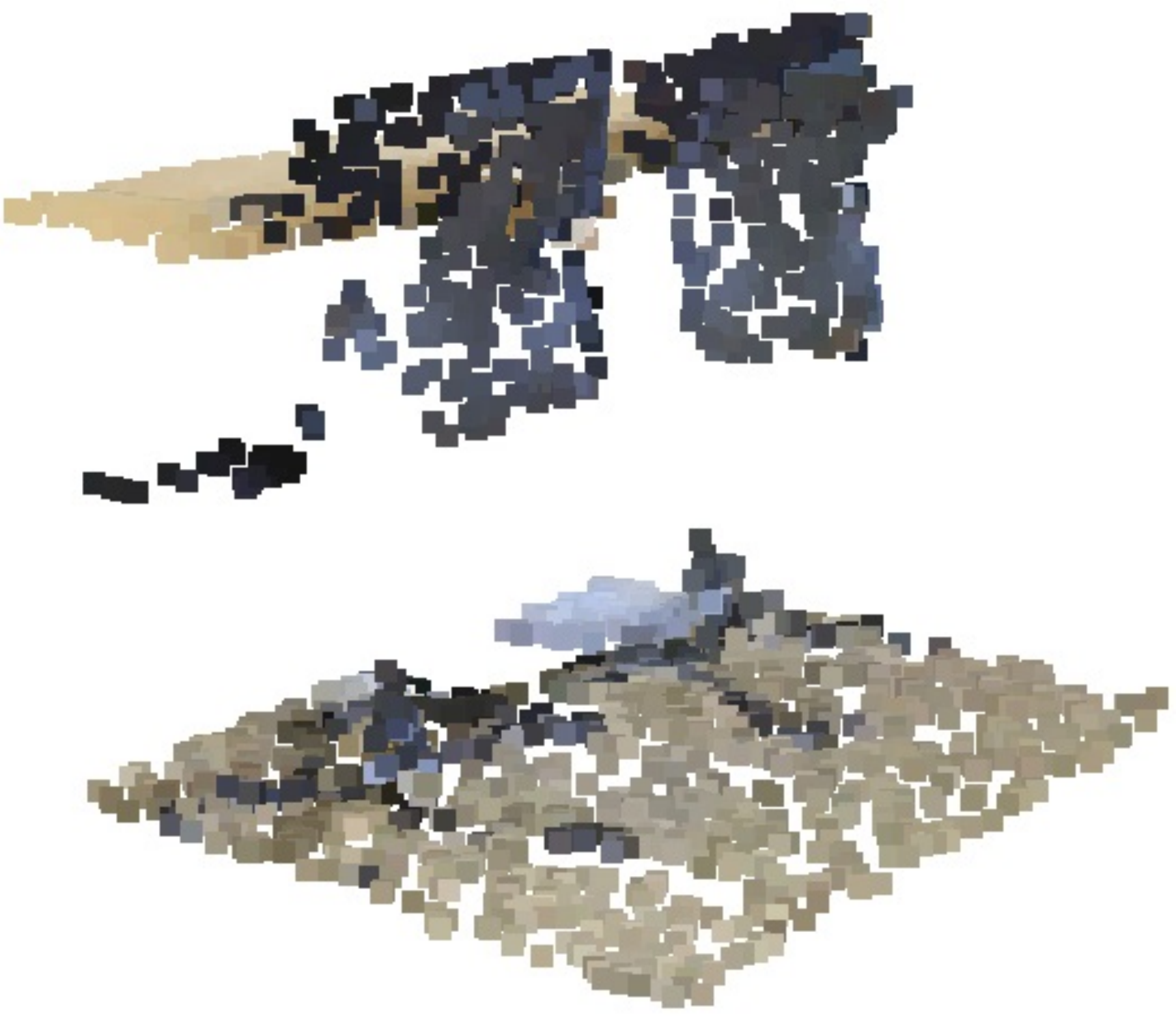}&
        \includegraphics[scale=0.141]{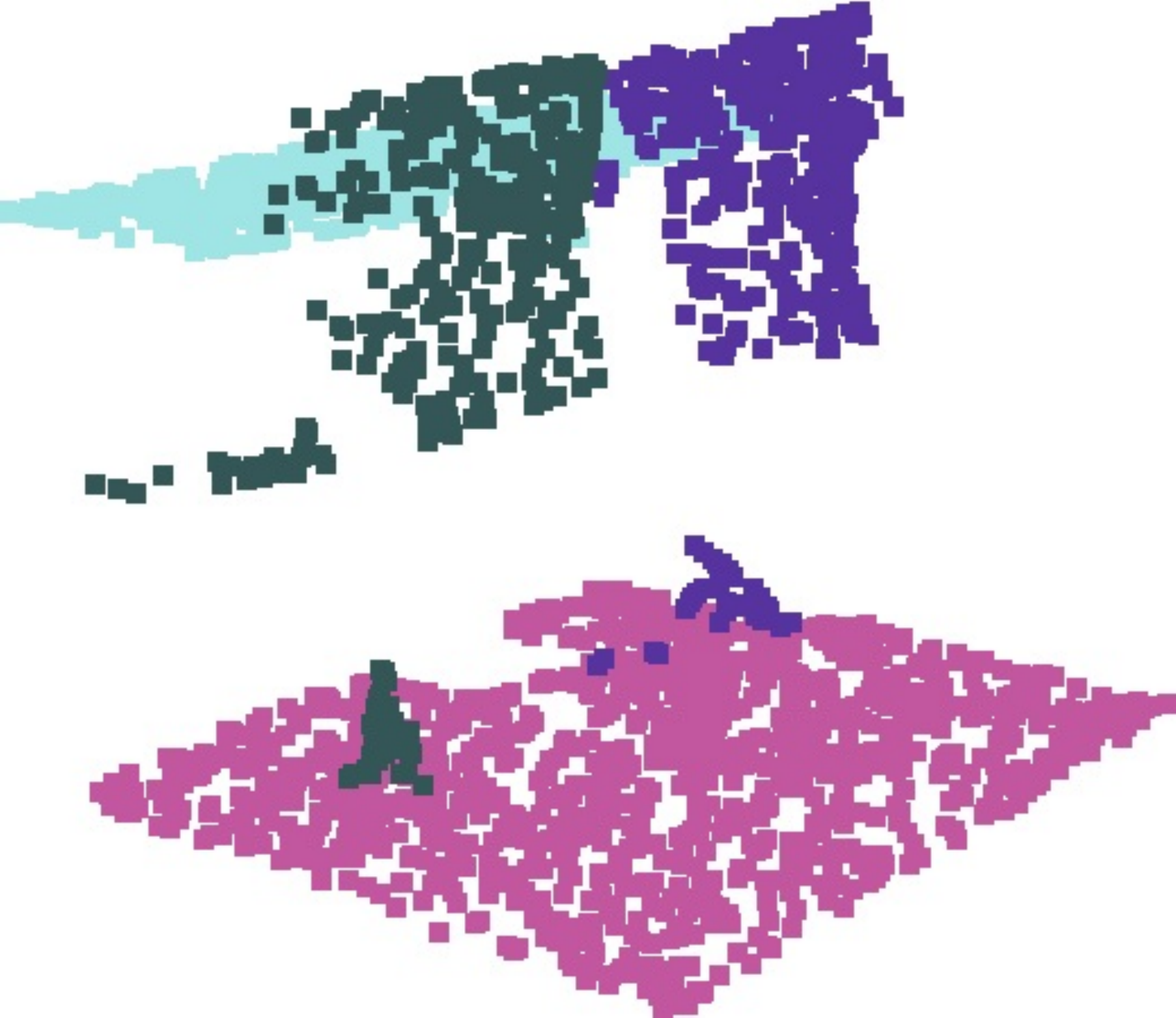}&
        \includegraphics[scale=0.141]{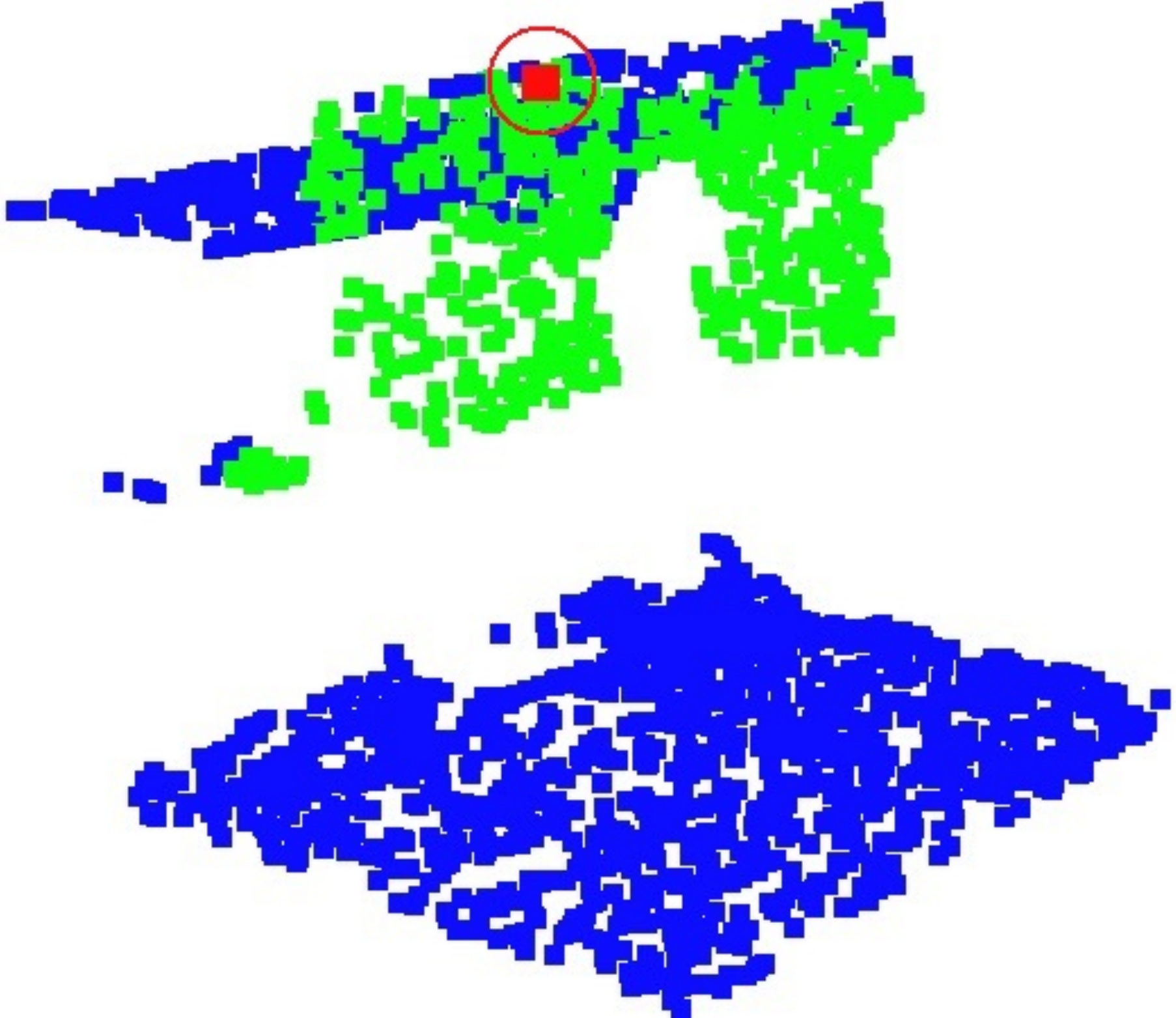}&
        \includegraphics[scale=0.141]{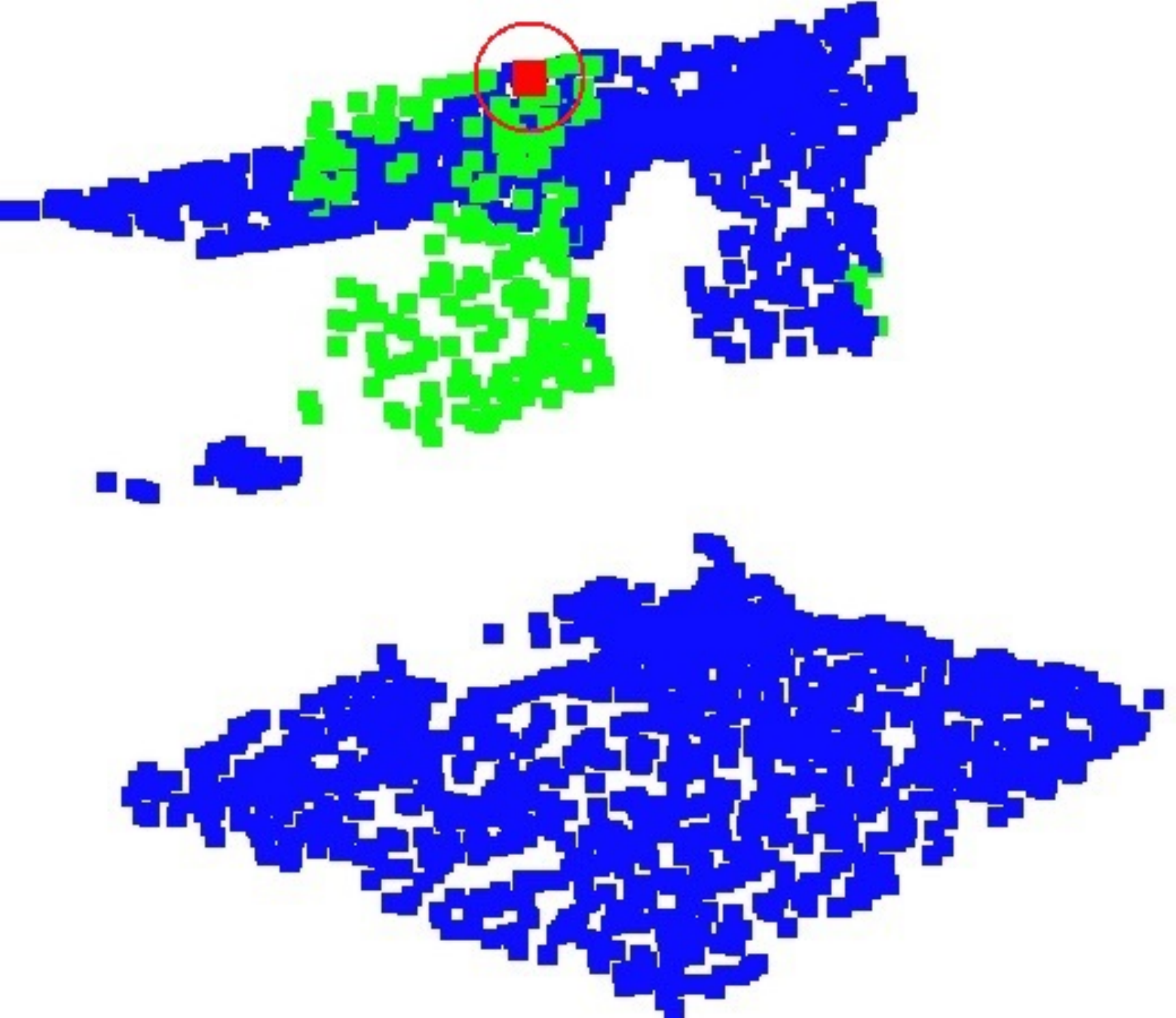}\\\\
        \includegraphics[scale=0.14]{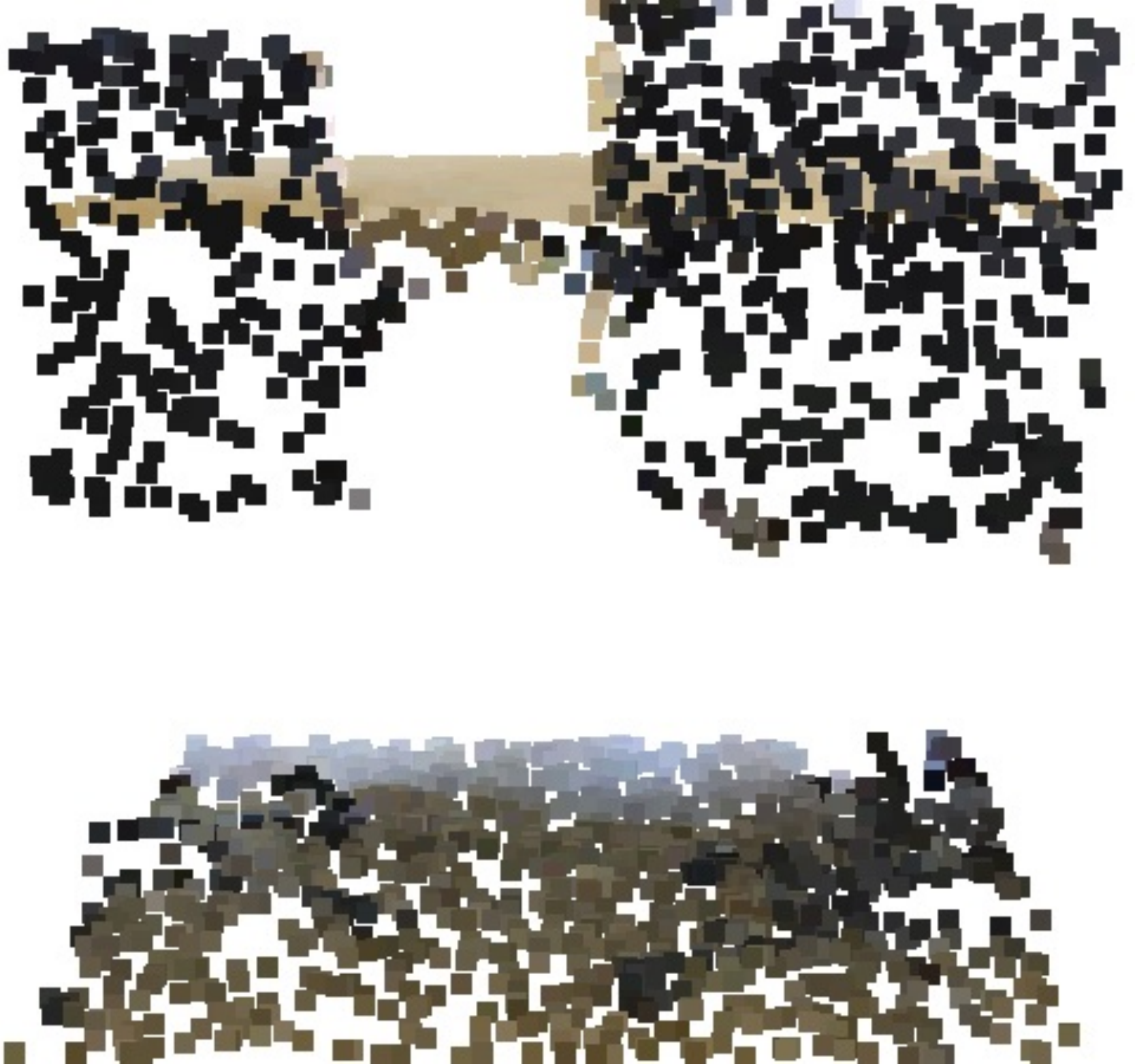}&
        \includegraphics[scale=0.14]{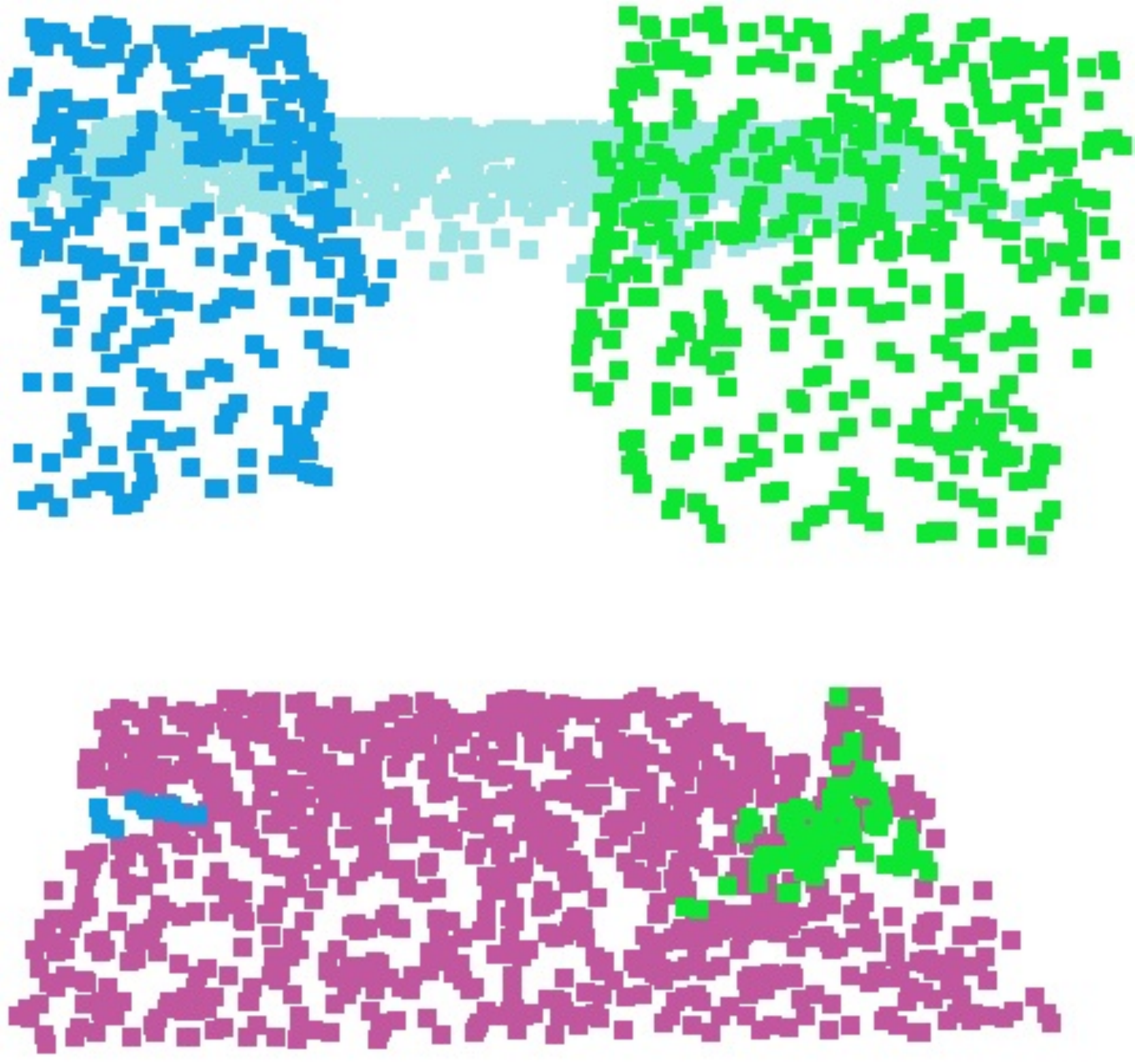}&
        \includegraphics[scale=0.14]{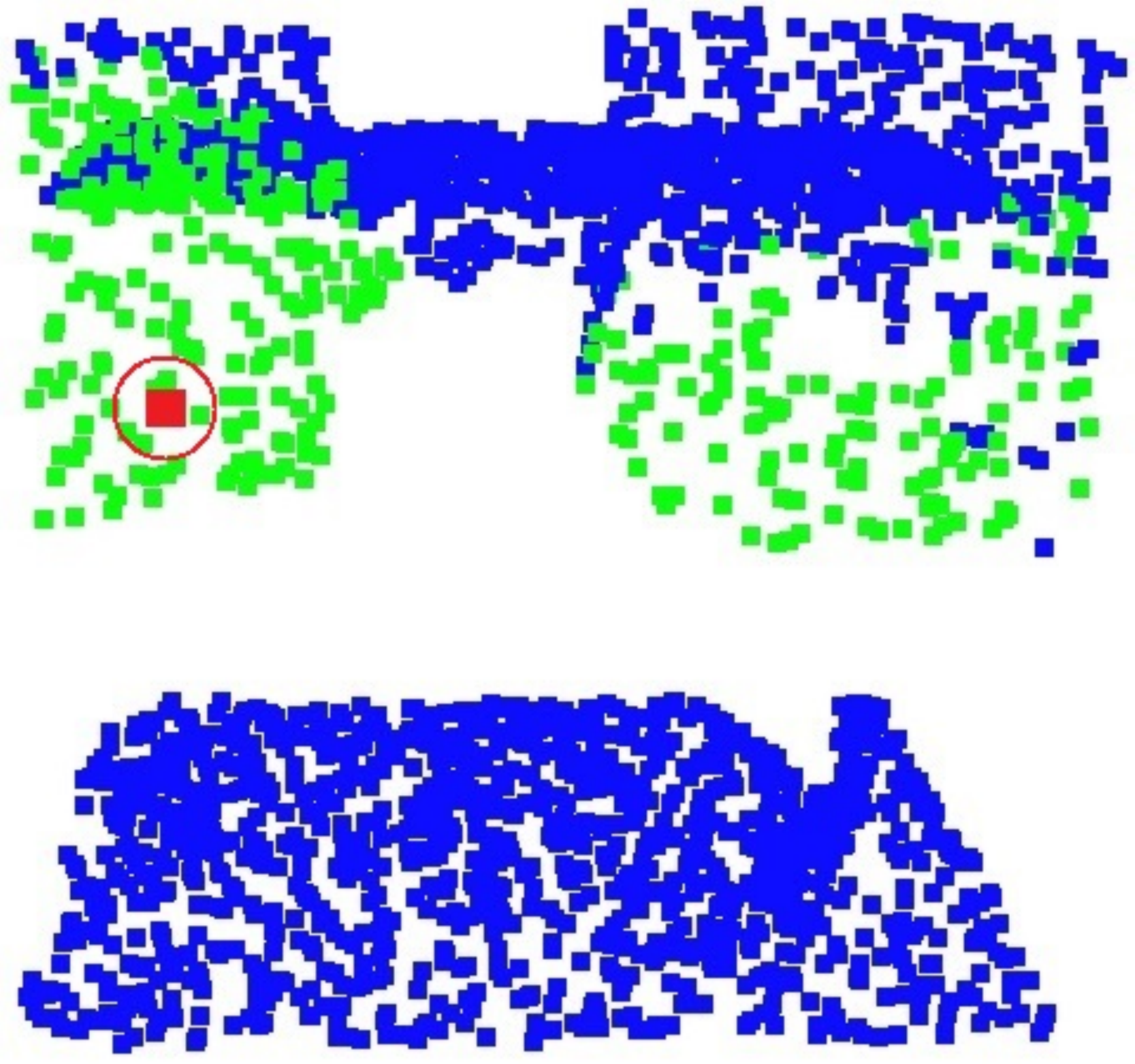}&
        \includegraphics[scale=0.14]{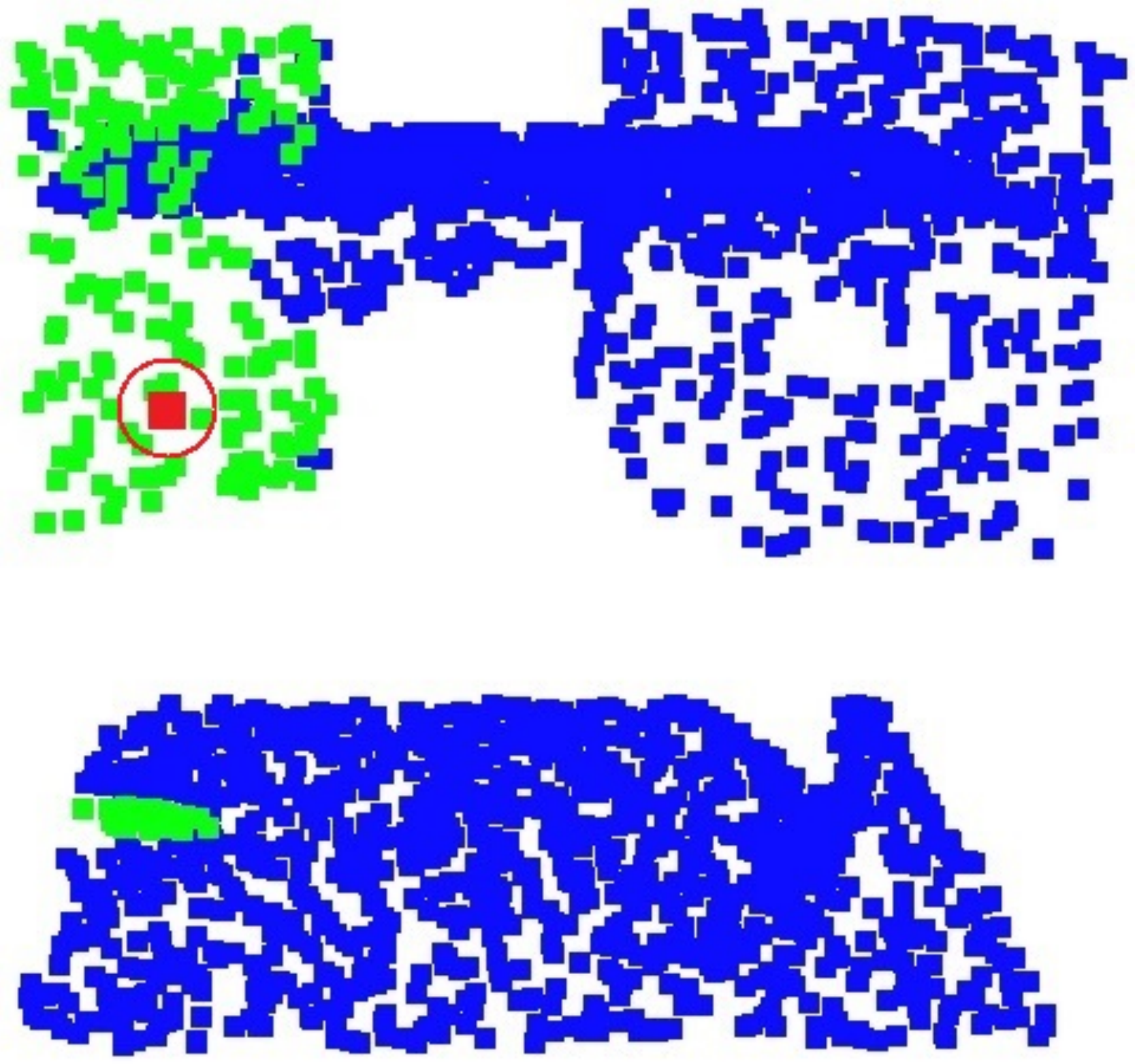}\\\\
        Real Scene&GT&Semantic Sim.&Instance Sim.\\
    \end{tabular}
    \caption{Visualization of instance and semantic similarity matrices. One row for each sample. From left to right, they are real scene blocks (each has two chairs), ground truth (instance), point cloud reflecting semantic similarity, point cloud reflecting instance similarity.}
    \label{fig:sim}
\end{figure}

In Fig.~\ref{fig:sim}, for trained networks and each sample, we select the same row from instance similarity matrix and semantic similarity matrix, respectively, then reshape the row vector to the 3D point cloud. So, the value of each point here represents the similarity to the point corresponding to the selected row. For better visualization, we binarize the 3D point cloud to divide points into two groups, similar points (green) and dissimilar points (blue) and marked the point corresponding to the selected row by red circle.
Each sample of Fig.~\ref{fig:sim} has two chairs in the scenes. We can see that the semantic similarity matrix could basically correctly reflect the semantic similarities, and the instance similarity matrix could highlight most of the points in the same instance.

\section{Conclusion}\label{sec:conclu}
We present Bi-Directional Attention Networks (BAN) for joint instance and semantic segmentation. Instead of element-wised fusing features for two tasks, our Bi-Directional Attention module builds instance and semantic similarity matrices from the instance and semantic features, respectively, with which two attention operations are conducted to bi-directionally aggregate features implicitly, introduce non-local information and avoid potential task conflict. Experiments on the S3DIS and PartNet datasets and method analysis suggest that the Bi-Directional Attention module could help give uniform and robust results within the same semantic or instance regions, and would also help to back-propagate uniform and robust gradients for optimization. Our BAN demonstrates significant superiority compared with baseline and other state-of-the-art works on the instance and semantic segmentation tasks consistently. Moreover, the ablation and mechanism study further verifies the design and effectiveness of the Bi-Directional Attention module.





%
%
\bibliographystyle{splncs04}
\bibliography{egbib}
\end{document}